\documentclass[lettersize,journal]{IEEEtran}

\usepackage[cmex10]{amsmath} 
\usepackage{amsfonts,stmaryrd,acronym,amssymb,amsthm,dsfont}
\usepackage{algorithm}
\usepackage{algorithmic}
\usepackage{graphicx,paralist,mdframed}
\usepackage{graphics}
\usepackage{bm}
\usepackage{url}
\usepackage{cite}
\usepackage{caption}
\usepackage{subcaption}
\usepackage{hyperref}
\usepackage{flushend}
\usepackage{enumerate}
\usepackage{pdflscape}
\usepackage[dvipsnames]{xcolor}
\usepackage{mathrsfs}
\usepackage{lipsum}
\usepackage{diagbox}
\usepackage{soul}
\usepackage{kantlipsum}
\usepackage{ragged2e} 
\usepackage{booktabs}

\usepackage[bbgreekl]{mathbbol}
\DeclareSymbolFontAlphabet{\mathbbm}{bbold}
\DeclareSymbolFontAlphabet{\mathbb}{AMSb}%

\graphicspath{{Graphics/}}
\definecolor{ColorHighlight}{rgb}{1,0,0}

\makeatletter
\def\blfootnote{\xdef\@thefnmark{}\@footnotetext}
\makeatother
\newcommand{\indi}[1]{\ensuremath{\mathds{1}}}

\newcounter{mytempeqcounter}

\theoremstyle{plain}

\theoremstyle{definition}

\theoremstyle{remark}
\newtheorem{remark}{Remark}

\allowdisplaybreaks
\allowbreak

\makeatletter%
\if@twocolumn%
\else
\fi%
\makeatother%

\newcommand{\indic}[1]{\ensuremath{\mathds{1}}}

\renewcommand{\leq}{\leqslant} 
\renewcommand{\geq}{\geqslant} 

\acrodef{ACDIS}[ACDIS]{Adaptive Communication Decision and Information Systems}
\acrodef{AEP}{Asymptotic Equipartition Property}
\acrodef{AWGN}{additive white gaussian noise}
\acrodef{AVC}[AVC]{Arbitrarily Varying Channel}
\acrodefplural{AVC}{Arbitrarily Varying Channels}
\acrodef{PIR-PNSI}{Private Information Retrieval with Private Noisy Side Information}
\acrodef{BER}{bit-error-rate}
\acrodef{BEC}{Binary Erasure Channel}
\acrodefplural{BEC}{Binary Erasure Channels}
\acrodef{BSC}{Binary Symmetric Channel}
\acrodefplural{BSC}{Binary Symmetric Channels}
\acrodef{BPSK}{Binary Phase-Shift Keying}
\acrodef{BICM}[BICM]{Bit-Interleaved Coded-Modulation}
\acrodef{CDF}[CDF]{Cumulative Distribution Function}
\acrodef{CGF}[CGF]{Cumulant Generating Function}
\acrodef{CLT}[CLT]{Central Limit Theorem}
\acrodef{DMC}[DMC]{Discrete Memoryless Channel}
\acrodefplural{DMC}{Discrete Memoryless Channels}
\acrodef{DMS}[DMS]{Discrete Memoryless Source}
\acrodef{ERM}[ERM]{Empirical Risk Minimization}
\acrodef{FER}[FER]{Frame Error Rate}
\acrodef{ICA}[ICA]{Independent Component Analysis}
\acrodef{iid}[i.i.d.]{independent and identically distributed}
\acrodef{IoT}[IoT]{Internet of Things}
\acrodef{KKT}[KKT]{Karush-Kuhn Tucker}
\acrodef{LASSO}[LASSO]{Least Absolute Shrinkage and Selection Operator}
\acrodef{LPD}[LPD]{Low Probability of Detection}
\acrodef{LDPC}[LDPC]{Low-Density Parity-Check}
\acrodef{CSCG}[CSCG]{Circularly Symmetric Complex Gaussian Distribution}
\acrodef{LLMS}[LLMS]{Linear Least Mean Square}
\acrodef{LMS}[LMS]{Least Mean Square}
\acrodef{MAC}[MAC]{Multiple-Access Channel}
\acrodef{ADSI}[ADSI]{Action-Dependent State Information}
\acrodef{MGF}[MGF]{Moment Generating Function}
\acrodef{MLC}[MLC]{multi-Level Coding}
\acrodef{MLE}[MLE]{maximum Likelihood Estimate}
\acrodef{MIMO}[MIMO]{multiple-Input Multiple-Output}
\acrodef{MISO}{multiple-Input Single-Output}
\acrodef{MSD}[MSD]{Multi-Stage Decoding}
\acrodef{MMSE}[MMSE]{minimum Mean-Square Error}
\acrodef{PAC}[PAC]{Probably Approximately Correct}
\acrodef{PCA}[PCA]{Principal Component Analysis}
\acrodef{PDF}[PDF]{Probability Density Function}
\acrodefplural{PDF}{Probability Density Functions}
\acrodef{PMF}[PMF]{Probability Mass Function}
\acrodefplural{PMF}{Probability Mass Functions}
\acrodef{PPM}[PPM]{Pulse Position Modulation}
\acrodef{PSD}{Power Spectral Density}
\acrodef{PSK}{Phase Shift Keying}
\acrodef{QKD}{Quantum Key Distribution}
\acrodef{ROC}{Receiver Operating Characteristic}
\acrodef{CVQKD}{Continuous-Variable \ac{QKD}}
\acrodef{QPSK}{Quadrature Phase-Shift Keying}
\acrodef{RV}{random variable}
\acrodef{SIMO}{Single-Input Multiple-Output}
\acrodef{SNR}{signal-to-noise ratio}
\acrodef{SVM}[SVM]{Support Vector Machine}
\acrodef{TPCP}{Trace-Preserving Completely-Positive}
\acrodef{wrt}[w.r.t.]{with respect to}
\acrodef{WSS}{Wide Sense Stationary}
\acrodef{RHS}{Right Hand Side}
\acrodef{LHS}{Left Hand Side}
\acrodef{PIR}{Private Information Retrieval}
\acrodef{MDS}{Maximum Distance Separable}
\acrodef{LLN}{law of Large Numbers}
\acrodef{DFRC}{dual-Function Radar Communication}
\acrodef{ISAC}{integrated sensing and communication}
\acrodef{RadCom}{Joint Radar and Communicatins}
\acrodef{PLS}[PLS]{Physical Layer Security}
\acrodef{RL}{reinforcement learning}
\acrodef{POCS}{projections onto convex sets}
\acrodef{SINR}{signal-to-interference-ratio}
\acrodef{RNN}{recurrent neural network}
\acrodef{BS}{base station}
\acrodef{MISO}{multiple-input-single-output}
\acrodef{MIMO}{multiple-input-multiple-output}
\acrodef{mmWave}{millimeter wave}
\acrodef{RF}{Radio frequency}
\acrodef{PS}{Phase shifter}
\acrodef{TTD}{true time delayer}
\acrodef{FDD}{frequency division duplex}
\acrodef{TDD}{time division duplex}
\acrodef{NN}{neural network}
\acrodef{CSI}{channel state information}
\acrodef{GAN}{generative Adversarial Network}
\acrodef{ULA}{uniform linear array}
\acrodef{BiCNN}{bi-directional convolutional NN}
\acrodef{EDN}{encoder-decoder network}
\acrodef{ISI}{inter-user interference}
\acrodef{KD-EDN}{knowledge-distillation-based EDN}
\acrodef{DNN}{deep neural network}
\acrodef{KD}{knowledge distillation}
\acrodef{AOA}{angle of arrival}
\acrodef{AOD}{angle of departure}
\acrodef{URLLC}{ultra-reliable low-latency communication}
\acrodef{L2O}{learning to optimize}
\acrodef{MRT}{maximum ratio transmission}
\acrodef{BRB}{branch-reduce-bound}
\acrodef{DL}{deep learning}
\acrodef{CNN}{convolutional neural network}
\acrodef{WMMSE}{weighted MMSE}
\acrodef{GNN}{graphical neural network}
\acrodef{IRS}{intelligent reflect surface}
\acrodef{CS}{compressed sensing}
\acrodef{SVD}{singular value decomposition}
\acrodef{ADMM}{alternating direction method of multipliers}
\acrodef{QoS}{quality of Service}
\acrodef{KDL}{KKT-guided dual learning}
\acrodef{PASS}{pinching antenna systems}
\acrodef{LSTM}{long short term memory}
\acrodef{SCA}{successive convex optimization}
\acrodef{RIS}{reconfigurable intelligent surface}
\acrodef{CL}{curriculum learning}
\acrodef{PE}{permutation equivariance}
\acrodef{MHSA}{multi-head self-attention}
\acrodef{MLP}{multi-layer perceptron}
\acrodef{PGA}{projected gradient ascent}
\acrodef{PGD}{projected gradient descent}
\acrodef{TBPTT}{truncated backpropagation through time}
\acrodef{SOTA}{state-of-the-art}
\acrodef{IUI}{inter-user interference}
\acrodef{PC}{positional coding}
\acrodef{FC}{fully-connected}
\acrodef{TN}{token-wise normalization}
\acrodef{HPE}{hierarchical permutation equivariance}
\acrodef{SGD}{stochastic gradient descent}
\acrodef{JRCB}{joint radar-communication beamforming}
\acrodef{RCS}{radar cross section}
\acrodef{LS}{least square}
\acrodef{SDP}{semidefinite program}
\acrodef{SAO}{semi-amortized learning-to-optimize}
\acrodef{SA-L2O}{semi-amortized learning-to-optimize}
\acrodef{MT-SAO}{masked-Transformer-based SAO}
\acrodef{MT-SAO-lift}{masked-Transformer-based SAO-lift}
\acrodef{DT-SAO-lift}{dedicated-Transformer-based SAO-lift}
\acrodef{SAO-ISAC}{SAO-for-ISAC}
\acrodef{LLM}{large language model}
\acrodef{SALLO}{semi-amortized lifted learning-to-optimize}
\acrodef{SALO}{semi-amortized learning-to-optimize}
\acrodef{SALLO-M}{semi-amortized lifted learning-to-optimize with masking}

\begin{document}

\title{Self-Evolving In-Context Learning for Direct Pilot-to-Beamformer Design in MU-MISO Systems}

\author{\IEEEauthorblockN{Yubo Zhang, and Xiaodong Wang}\\
\thanks{Y.~Zhang and X.~Wang are with the Department of Electrical Engineering, Columbia University, New York, NY 10027.}}

\maketitle
\thispagestyle{empty}

\begin{abstract}
We develop an enhanced in-context learning (ICL) framework to improve the performance of pilot-based beamforming 
in multi-user multiple-input single-output (MU-MISO) systems.
The proposed scheme integrates the ICL-Transformer backbone with the pilot encoder-decoder network (EDN) and the beamformer EDN. A crucial feature of our ICL network is that it can handle multiple channel models without retraining, enabled by the construction of model-specific context datasets. To improve convergence and robustness, we introduce three key innovations: (a) a curriculum learning (CL) strategy that smoothly transitions from supervised LMMSE-labeled imitation to unsupervised sum-rate maximization, (b) a self-evolving mechanism that dynamically expands and refines the context datasets for all channel models during CL-based training, and (c) a mismatch-aware extension that incorporates several mismatches into the general ICL framework and bypasses explicit channel calibrations. Ablation studies validate the effectiveness of the in-context architecture and enhanced training strategies. Simulation results over diverse communication environments show that the proposed scheme is able to rapidly adapt to both seen and unseen channel models without gradient-based parameter updates, and can mitigate the mismatch issues via intelligent context constructions. Furthermore, our scheme consistently outperforms the existing beamforming schemes under pilot-based settings, including the WMMSE benchmark and the recent Transformer-based methods.



\end{abstract}

\begin{IEEEkeywords}
In-context learning, pilot-based beamforming, self-evolving, curriculum learning, multi-model adaptation, mismatch-aware design
\end{IEEEkeywords}


\section{Introduction}

Next-generation wireless systems are expected to operate with larger antenna arrays, denser user populations, and more diverse propagation environments, creating an urgent need for scalable, generalizable and low-latency physical-layer signal processing~\cite{bjornson2014optimal}. 
Classical iterative methods such as the weighted minimum mean squared error (WMMSE) algorithm~\cite{shi2011wmmse} and the branch-reduce-and-bound (BRB) method~\cite{Bjornson2013} can attain near-optimal solutions, but their computational cost is often prohibitive in large-scale systems. At the other extreme, low-complexity linear schemes such as maximum-ratio transmission (MRT)~\cite{Lozano2006} and linear MMSE (LMMSE) beamforming~\cite{bjornson2014optimal} provide fast solutions at the cost of substantial performance degradation. To bridge this complexity-performance gap, \ac{DL} approaches, especially Transformer-based models, have recently been investigated to enable high-quality beamforming with improved scalability and real-time adaptability.

\subsection{Pilot-based Beamforming Learning}

Deep learning for beamforming and precoding in multi-user systems has undergone rapid development in recent years. Early works mainly relied on lightweight architectures such as fully-connected neural networks (FCNNs)~\cite{xia2020deepframework,zhang2026semi,zhang2025encoder,zhang2025power} and convolutional neural networks (CNNs)~\cite{sohrabi2021dl_feedback,zhang2026direct,zhang2024channel} to learn compact channel representations or to directly predict beamformers. More recently, graph neural networks (GNNs)~\cite{kim2022bipartite_gnn,li2024GNN} have been introduced to improve scalability by exploiting the interaction structures among users and antennas, while Transformer-based schemes~\cite{duan2025learning_precoding_tf,zhang2024cnn_tf,zhang2025sallom} further enhance the expressive power by capturing long-range global dependencies in large-scale optimization problems. In parallel, several studies have begun to explore wireless foundation models~\cite{wen2026wifo,xu2025llmnearfield} toward broader multi-task and multi-scenario adaptability.

Despite these advances, most learning-based beamforming methods still assume perfect CSI at the base station (BS), whereas practical systems provide only noisy, limited-length pilot signals. In particular, when the pilot length is smaller than the number of users, the full channel cannot be exactly recovered even in the noiseless case~\cite{park2024e2e_tdd,zhang2022bi}. This information bottleneck limits the conventional two-stage pipeline of channel estimation followed by beamforming~\cite{jang2022joint_pilot_feedback,ding2023266}: channel estimation minimizes reconstruction error rather than the ultimate sum-rate objective, and its errors propagate to the subsequent beamforming stage. These limitations motivate \emph{direct pilot-to-beamformer learning}, which generates beamformers directly from pilot observations without explicitly reconstructing the full channel~\cite{sohrabi2021dl_feedback,attiah2022dl_sensing}. This approach is particularly well suited to sparse mmWave and sub-THz channels, where angular sparsity enables compressed pilots to retain beamforming-relevant information~\cite{jang2022joint_pilot_feedback,li2019auto_precoder}. Accordingly, encoder-decoder network (EDN) architectures can extract task-relevant features from the pilots and map them directly to beamforming solutions, an approach whose effectiveness has been demonstrated in recent studies~\cite{park2024e2e_tdd,zhang2026semi,pan2025beacon}.

\subsection{In-Context Learning}

In-context learning (ICL) has emerged as a promising paradigm for multi-scenario adaptation without parameter updates~\cite{dong2022icl_survey,pan2024beacon}. Originally popularized by large language models, ICL enables a model to infer the underlying task from a few input demonstrations and adapt through forward inference alone~\cite{wies2023learnability}. Theoretical studies have further shown that self-attention can implicitly emulate gradient-based adaptation, providing a principled interpretation of this capability~\cite{garg2022what,vonoswald2023icl_gd}.

In wireless communications, ICL has been applied primarily to receiver-side detection and estimation, where pilot-based demonstrations enable real-time adaptation without retraining~\cite{zecchin2024cellfree,zecchin2025icl_receiver}. Recent results further establish Transformer-based ICL as an optimal in-context estimator for certain wireless estimation problems~\cite{kunde2025icl_wireless}. Of particular relevance to our work, \cite{wen2026icwlm} extends ICL to transmitter-side beamforming using demonstration pairs for task adaptation. Nevertheless, it assumes perfect CSI and requires a large corpus of pre-computed WMMSE labels, limiting its practicality for large-scale high-frequency systems.


\subsection{Curriculum Learning and Self-Evolving Mechanism}

Our training framework combines curriculum learning (CL) with a self-evolving mechanism. CL gradually increases the learning difficulty~\cite{bengio2009curriculum,soviany2022curriculum_survey}, facilitating a stable optimization and reducing the risk of poor early-stage convergence. This is particularly beneficial for beamforming, whose non-convex sum-rate objective becomes increasingly difficult under heavier user loads and larger system scales. Recent studies~\cite{johnston2023curriculum_l2o,zhang2025sallom} have also demonstrated that CL can improve the scalability and generalization of learning-based beamforming methods.

Complementing CL, the self-evolving strategy progressively augments a small seed dataset with high-quality model-generated samples. This idea is related to self-training and pseudo-labeling~\cite{lee2013pseudo_label,amini2025selftraining_survey}, where carefully designed quality control prevents unreliable predictions from degrading training~\cite{xie2020noisy_student}. Similar iterative learning from filtered model generations has also shown promise for large language models~\cite{chen2024spin}. In our ICL framework, this mechanism reduces the reliance on expensive pre-computed beamforming labels while continuously enriching the context distribution. More importantly, the resulting context dataset provides increasingly diverse and informative demonstrations, thereby improving the model's generalization and adaptation capabilities.

\subsection{Contributions and Outline}

Despite recent progress, existing pilot-based beamforming methods remain limited in several respects. Conventional pilot-to-beamformer schemes typically employ conventional neural networks tailored to fixed channel models, resulting in limited scalability and cross-scenario generalization~\cite{li2019auto_precoder,sohrabi2021dl_feedback,attiah2022dl_sensing,park2024e2e_tdd}. Transformer-based schemes improve scalability, but generally remain scenario-specific~\cite{duan2025learning_precoding_tf,zhang2024cnn_tf}; extending them to heterogeneous channel environments often requires additional modules~\cite{wen2026wifo}, architectural modifications, or costly retraining and fine-tuning~\cite{zhang2025sallom}. Moreover, the most relevant ICL-based beamforming method, ICWLM~\cite{wen2026icwlm}, assumes perfect CSI and requires a large corpus of pre-computed WMMSE labels, limiting its applicability to practical pilot-based high-frequency systems. To tackle these issues, we propose a self-evolving ICL framework with context bootstrapping for direct pilot-to-beamformer design in MU-MISO systems, enabling scalable multi-model adaptation from limited-length pilots with substantially fewer pre-computed labels. Our main contributions are summarized as follows.
\begin{itemize}
    \item \textbf{ICL-Transformer architecture:}
    We develop an enhanced ICL framework that directly generates beamformers from noisy, length-limited pilot signals without explicit channel estimation. The proposed architecture integrates pilot EDN and beamformer EDN with an ICL Transformer backbone, allowing the Transformer to reason based on the pilot-beamformer demonstration pairs in a low-dimensional feature space.

    \item \textbf{Curriculum self-evolving training with context bootstrapping:}
    We design a CL-based training scheme that gradually transitions from supervised imitation of inexpensive LMMSE labels to unsupervised sum-rate maximization. This strategy prevents the early-stage poor convergence to local optima and enables the model to improve beyond the initial solutions without requiring near-optimal labels. To improve the training effect, a self-evolving strategy that dynamically expands and refines the context datasets is adopted, where model-generated solutions are selectively admitted into the datasets. This mechanism reduces the dependence on pre-computed labels while providing increasingly diverse and informative demonstrations for ICL.


    \item \textbf{Multi-model adaptation via context datasets:}
    A single shared ICL network can adapt to multiple channel models through model-specific context datasets, without adding additional modules or updating network parameters. We further extend its adaptation space to model-mismatch settings by incorporating several mismatches into the proposed ICL framework and bypassing explicit channel calibrations.
    
    \item \textbf{Performance evaluations:}
    Ablation studies justify the effectiveness of the the proposed ICL-Transformer architecture and enhanced training strategies. Extensive simulations demonstrate that the proposed scheme rapidly adapts to both seen and unseen channel models without gradient-based parameter updates, while effectively mitigating model mismatches through context construction. Furthermore, our scheme consistently outperforms the existing beamforming schemes under pilot-based settings, including WMMSE benchmark and recent Transformer-based methods.
\end{itemize}




\section{Background} \label{bg_intro}

\subsection{Downlink Beamforming} \label{bf_prob_intro}

Consider a TDD MU-MISO downlink system, where a base station (BS) equipped with $N$ transmit antennas serves $K$ single-antenna users. The downlink channel from the BS to user $k$ is denoted by $\bm{h}_k \in \mathbb{C}^{N \times 1}$, and the aggregated channel matrix is $\bm{H} = [\bm{h}_1, \ldots, \bm{h}_K] \in \mathbb{C}^{N \times K}$. 
Assume that the BS applies a linear beamformer $\bm W=[\bm w_1,\ldots,\bm w_K]\in\mathbb{C}^{N\times K}$ to transmit user symbols $\bm a=[a_1,\ldots,a_K]^T$, where $\mathbb{E}[\|a_k\|^2]=1$ for all $k$. Then the transmitted signal is given by
\begin{align}
\bm x=\sum_{i=1}^{K}\bm w_i a_i=\bm W \bm a.
\end{align}
Accordingly, the received signal at user $k$ is
\begin{align}
r_k=\bm h_k^H\bm w_k a_k+\sum_{i\neq k}\bm h_k^H\bm w_i a_i+z_k,
\end{align}
where $z_k\sim\mathcal{CN}(0,\sigma_k^2)$ denotes the additive noise at user $k$. 
Hence the achievable sum-rate is
\begin{align} \label{sum_rate_def}
R_{\rm sum}(\bm H,\bm W)
=
\sum_{k=1}^{K}
\log_2\!\left(
1+\frac{|\bm h_k^H\bm w_k|^2}
{\sum_{i\neq k}|\bm h_k^H\bm w_i|^2+\sigma_k^2}
\right).
\end{align}
Given the channel state information (CSI) $\bm{H}$, the optimal beamformer is then formulated as 
\begin{align} \label{opt_bf_map_def}
\bm{W}^*(\bm{H}) = \arg\max_{\|\bm{W}\|_F^2 \leq P} R_\text{sum}(\bm{H},\bm{W}),
\end{align}
where $P$ is the maximum transmit power budget at the BS. Typically, the optimal beamformer in \eqref{opt_bf_map_def} is hard to compute. In contrast, the suboptimal LMMSE beamformer $\bm{W}^{\text{m}}(\bm{H}) = [\bm{w}^{\text{m}}_1,\dots,\bm{w}^{\text{m}}_K]$ is much easier to obtain, given by
\begin{align} \label{mmse_bf}
\bm{w}^{\text{m}}_k = \sqrt{\frac{P}{K}} \cdot \frac{ (\bm{I}_N + \sum_{i=1}^{K} \frac{P}{K} \bm{h}_i \bm{h}_i^H)^{-1} \bm{h}_k }{ \Vert (\bm{I}_N + \sum_{i=1}^{K} \frac{P}{K} \bm{h}_i \bm{h}_i^H)^{-1} \bm{h}_k \Vert_2}, \ k \in [K].
\end{align}


In practice, the CSI is not available at the BS. Instead, the BS only has access to the received uplink pilot signal. Specifically, let $\bm{\Psi} \in \mathbb C^{K\times L_p}$ denote the pilot matrix, where $L_p$ is the pilot length. The received pilot signal at the BS is given by
\begin{align} \label{pilot_sig_model}
\bm Y=\bm H \bm\Psi+\bm N \in \mathbb{C}^{N\times L_p},
\end{align}
where $\bm N\in\mathbb{C}^{N\times L_p}$ denotes the additive white Gaussian noise with i.i.d. entries $n_{ij}\sim\mathcal{CN}(0,\sigma_B^2)$. Since pilot transmission consumes valuable time-frequency resources, the pilot length is usually limited, especially in large-scale high-frequency systems. Therefore, communication systems typically operate in the underdetermined regime such that $L_p<K$, where the channel matrix cannot be uniquely recovered even in the noiseless case, since any perturbation $\Delta \boldsymbol H$ satisfying $\Delta \bm H \cdot \bm\Psi=\bm{0}$ is unobservable from the pilot signal.

Several learning-based approaches to beamforming optimization~\cite{duan2025learning_precoding_tf,zhang2024cnn_tf,zhang2025sallom}
have been proposed to approximate the optimal beamformer in \eqref{opt_bf_map_def}, assuming that perfect CSI is available at the BS. When only the pilot signal
is available, existing learning-based methods~\cite{Ma2020Sparse,Elbir2022Family} typically follow a two-stage process: first an estimate $\hat{\bm{H}}$ of the channel matrix is obtained based on the pilot signal $\bm{Y}$, and then the beamformer is formed using the estimated CSI $\hat{\bm{H}}$. Moreover, recent works~\cite{sohrabi2021dl_feedback,attiah2022dl_sensing} advocate \emph{direct pilot-to-beamformer} learning, where the received pilot signal is mapped directly to the beamformer without explicit channel estimation. Such an approach can potentially avoid error propagation from the imperfect CSI recovery and use the short pilot signal more efficiently.




\subsection{In-Context Learning} \label{icl_bg_intro}


We briefly review the traditional in-context learning (ICL) mechanism and its implementation based on  Transformer models. When a single model is expected to operate across \emph{multiple tasks or scenarios}, a central challenge lies in the rapid adaptation to the specific scenario without explicit supervision or retraining. ICL addresses this challenge by presenting a few input-output samples as part of the model input, which serve as the context and allow the model to infer the task/scenario information directly from the input sequence, rather than relying on specific architectural branches or explicit metadata. Specifically, denote by $\mathcal{S}$ the set of all scenarios. For each scenario $s \in \mathcal{S}$, an $\ell$-shot input sequence to the neural network model $\mathcal{M}_{\bm{\theta}}$ is formed as 
\begin{equation} \label{ell_shot_sequence}
\mathcal{Z}_s^\ell = [(\bm{z}_{1}, f(\bm{z}_{1})), (\bm{z}_{2}, f(\bm{z}_{2})), \ldots, (\bm{z}_{\ell}, f(\bm{z}_{\ell})), \bm{z}_{\ell+1}],
\end{equation}
where $\mathcal{C}_s = [(\bm{z}_{j}, f(\bm{z}_{j}))]_{j=1}^\ell$ denotes the input-output samples, referred to as the \emph{context}, $\bm{z}_{\ell+1}$ denotes the \emph{query input}, and $f(\cdot)$ denotes the objective function. The model then generates the \emph{query output} $\mathcal{M}_{\bm{\theta}}(\mathcal{Z}_s^\ell)$ to predict the ground truth $f(\bm{z}_{\ell+1})$. 

During training, the model acquires its ICL capability
by minimizing the following expected loss over all scenarios $s \in \mathcal{S}$ and over all input-output pairs:
\begin{equation} \label{icl_mt_pretrain}
\min_{\bm{\theta}} \ \sum_{s \in \mathcal{S}} \mathbb{E}_{\mathcal{Z}_s^\ell} \bigg[ \mathcal{L}\big(\mathcal{M}_{\bm{\theta}}(\mathcal{Z}_s^\ell), f(\bm{z}_{\ell+1})\big) \bigg],
\end{equation}
where $\mathcal{L}(\cdot,\cdot)$ is a properly chosen loss function, such as the mean squared error (MSE) for regression tasks.

Transformer architectures are particularly suitable for ICL frameworks~\cite{garg2022what,vonoswald2023icl_gd}. This is because ICL naturally organizes the input data as a context-query token sequence, where each token represents either an input or an output sample. Owing to its strong sequence-modeling capability, Transformer can effectively process such a context-query sequence and serves as a query-output predictor. In the context of wireless communication applications, recent works~\cite{zecchin2025icl_receiver,zecchin2024cellfree} show that ICL Transformers are well suited for receiver-side functionalities, such as symbol detection. Furthermore, \cite{wen2026icwlm} demonstrates the potential of ICL Transformers for transmitter-side optimization, based on perfect CSI and a large labeled dataset.




\section{Pilot-based Beamforming} \label{pilot_bf_scheme}

\subsection{Problem Statement} \label{prob_state}

We aim to train a single neural network that handles direct pilot-to-beamformer learning across multiple channel models. Let $\mathcal{S}$ denote the set of all channel models, where each $s\in\mathcal{S}$ corresponds to a channel distribution $p_s$, i.e., $\bm{H} \sim p_s$. As discussed in Sec.~\ref{bf_prob_intro}, for a given channel model $s \in \mathcal{S}$, a \textbf{CSI-based} beamforming optimization scheme learns a mapping $\bm{W}(\bm{H})$ that approaches the optimal mapping $\bm{W}^*(\bm{H})$ in \eqref{opt_bf_map_def}, for $\bm{H} \sim p_s$. On the other hand, to have a single network for multiple channel models, the ICL can be employed~\cite{wen2026icwlm}. In particular, for each channel model $s \in \mathcal{S}$, a context dataset $\mathcal{V}_s$ composed of samples $(\bm{H},\bm{W}^*(\bm{H}))$ can be precomputed, where $\bm{H} \sim p_s$, and $\bm{W}^*(\bm{H})$ is the corresponding near-optimal WMMSE beamformer. The network input includes a context set $\mathcal{C}_s$ consisting of $\ell$ samples from $\mathcal{V}_s$, and a new channel $\bm{H} \sim p_s$, and the expected output is the corresponding beamformer $\bm{W}^*(\bm{H})$. However, such an approach requires computing a large number of near-optimal beamformer labels, which is computationally expensive, and becomes infeasible for large-scale MIMO systems. 

\begin{remark}
Meta-learning~\cite{Finn2017MAML} is an alternative approach to achieve the fast adaptation to multiple channel models. However, meta-learning requires inner-loop and outer-loop gradient updates for task adaptations, while ICL keeps parameters of the pretrained model fixed and performs gradient-free adaptations via context. Hence the ICL approach is selected in this work owing to a more lightweight online adaptation.
\end{remark}

In this paper, our goal is to learn a direct mapping from the pilot signal to the beamformer across different channel models. Note that such a \textbf{pilot-based} beamforming cannot be formulated similar to \eqref{opt_bf_map_def} with $\bm H$ simply replaced by $\bm Y$, since unlike \eqref{sum_rate_def}, there is no explicit expression for the sum rate based on the pilot signal $\bm{Y}$.

Nevertheless, a straightforward way is to simply modify the above ICL approach as follows: for each channel model $s\in\mathcal{S}$, we draw channel samples $\bm{H} \sim p_s$, and form the corresponding pilot signal $\bm{Y}$ according to \eqref{pilot_sig_model} and compute the WMMSE beamformer $\bm{W}^*(\bm{H})$. The context dataset $\mathcal{V}_s$ is then composed of samples $(\bm{Y},\bm{W}^*(\bm{H}))$. However, such an approach suffers from the same prohibitively high complexity for precomputing the context datasets for large-scale systems.

To that end, we propose an \textbf{enhanced ICL scheme} based on the curriculum learning (CL) paradigm, in which the model first learns from low-complexity LMMSE labels to obtain a stable initialization, and then gradually shifts toward unsupervised sum-rate maximization on query samples. In particular, the context datasets $\mathcal{V}_s$, $s\in\mathcal{S}$, are not precomputed, but expanded and refined during the training stage. Moreover, during training only LMMSE beamformers are computed and no WMMSE beamformer is needed.

Specifically, during training, for each channel model $s \in \mathcal{S}$, we dynamically update a context dataset $\mathcal{V}_s$ consisting of triplets $(\bm{H},\bm{Y},\hat{\bm{W}})$. For a channel sample $\bm{H} \sim p_s$, the pilot signal $\bm{Y}$ is generated according to \eqref{pilot_sig_model}, and the beamformer solution $\hat{\bm{W}}$ is either an LMMSE beamformer computed by \eqref{mmse_bf}, or a model-generated beamformer. In particular, initially, we set 
\begin{align} \label{ini_dataset_def}
\mathcal{V}^{(0)} = \bigcup_{s \in \mathcal{S}} \mathcal{V}_s^{(0)}, \ \mathcal{V}_s^{(0)} = \bigl\{(\bm{H}_{j}, \bm{Y}_{j}, \bm{W}_{j}^{\text{m}})\bigr\}_{j=1}^{M_0},
\end{align}
where $\boldsymbol Y_{j}$ and $\boldsymbol W^{\text{m}}_{j}$ are the corresponding pilot signal and LMMSE beamformer based on the channel sample $\bm{H}_j \sim p_s$. Following the ICL formulation in \eqref{ell_shot_sequence} and \eqref{icl_mt_pretrain}, the context is formed by randomly selecting $\ell$ pilot-beamformer demonstration pairs from $\mathcal{V}_s$:
\begin{align} \label{context_cs_def}
\mathcal{C}_s = [(\bm{Y}_j, \hat{\bm{W}}_j)]_{j=1}^{\ell}.
\end{align}
Given a new query input $\bm{Y}$, the context-query sequence is written as $\mathcal{Z}_s^\ell = [\mathcal{C}_s,\bm{Y}]$. Then the context-conditioned pilot-to-beamformer mapping can be represented by a neural network $\mathcal{M}_{\bm{\theta}}$:
\begin{align} \label{icl_forward_demo}
\bm{W} = \mathcal{M}_{\bm{\theta}}(\mathcal{Z}_s^\ell).
\end{align}
Then the ultimate objective is to maximize the expected sum-rate over all channel models:
\begin{align} \label{amor_icl_formu}
\max_{\bm{\theta}} \ \sum_{s \in \mathcal{S}} \
\mathbb E_{\substack{ \boldsymbol H \sim p_s, \bm{N} \\ \boldsymbol Y=\boldsymbol H \boldsymbol\Psi+\boldsymbol N\\ \mathcal C_s \in \mathcal{V}_s(\bm{\theta})}}
\left[R_\text{sum} \left(\bm{H}, \mathcal{M}_{\bm{\theta}}(\mathcal{Z}_s^\ell)\right)\right].
\end{align}
Note that the context dataset $\mathcal{V}_s$ depends on the network parameters $\bm{\theta}$ and therefore is dynamically updated during training, as will be detailed in Sec.~\ref{evolve_icl_framework}. Starting from these initial datasets, we adopt a CL-based training scheme that combines a supervised MSE loss with an unsupervised sum-rate loss as the objective. At the same time, starting from $\mathcal{V}_s^{(0)}$, each context dataset $\mathcal{V}_s$ is expanded and updated to improve its quality. Consequently, the training procedure produces both a trained ICL model and a context dataset $\mathcal{V}_s$ for each channel model $s \in \mathcal{S}$.

\emph{During inference}, given the received pilot signal $\bm{Y}$ and its underlying channel model $s$, a context $\mathcal{C}_s$ is sampled from the dataset $\mathcal{V}_s$, and the beamformer solution is generated by a single forward pass $\bm{W} = \mathcal{M}_{\bm{\theta}^*}(\mathcal{Z}_s^\ell)$, where $\mathcal{Z}_s^\ell = [\mathcal{C}_s,\bm{Y}]$, and $\mathcal{M}_{\bm{\theta}^*}$ is the trained model. 

\begin{remark}
Our proposed enhanced ICL scheme differs from existing ICL-based schemes~\cite{wen2026icwlm,zecchin2025icl_receiver,kunde2025icl_wireless}: In those prior works, the demonstration pairs and the query pair are typically assumed to follow the same deterministic input-output rule, and the context dataset is precomputed and fixed during both training and inference. 
\end{remark}



\subsection{Overall Network Architecture} \label{net_arch_intro}

\begin{figure*}
    \centering
    \includegraphics[width=1.02\linewidth]{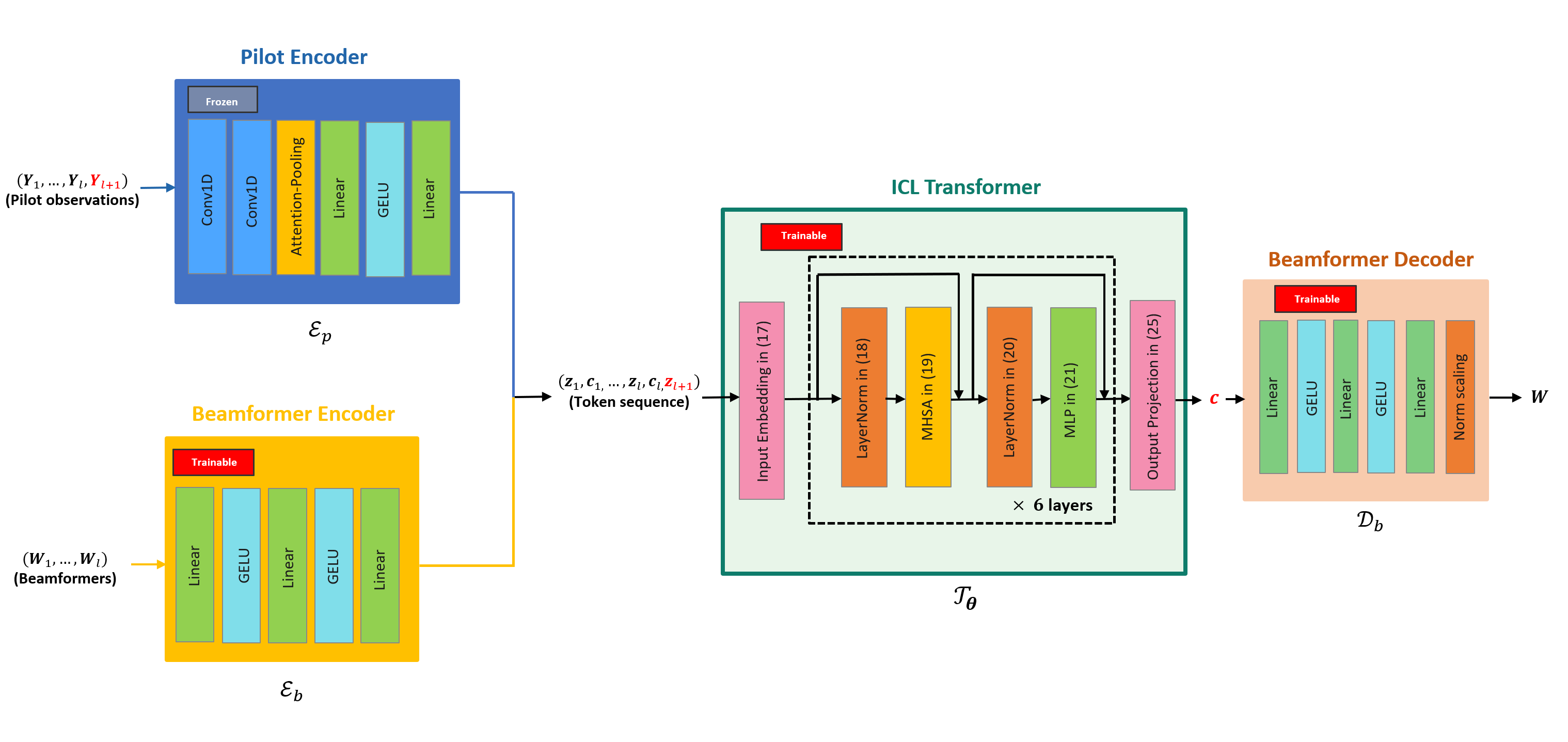}
    \caption{The network architecture of the proposed ICL beamforming framework. }
    \label{icl_net_arch}
\end{figure*}

We design a network architecture tailored to the ICL beamformer learning problem described in Sec.~\ref{prob_state}, as illustrated in Fig.~\ref{icl_net_arch}. To effectively capture the features of context samples, a pretrained pilot encoder network and a pretrained beamformer encoder network are adopted to compress the pilot signals and beamformers into the same-size low-dimensional tokens, respectively. This also reduces the subsequent computation costs and enables efficient context-query processing. The resulting compressed token sequence is then fed into a Transformer backbone, and finally the beamformer solution is generated by a beamformer decoder network. 



\subsection{EDN Pretraining} \label{edn_pretrain_sec}


We use a convolutional encoder network with attention pooling to compress the pilot signal, aiming to
exploit its structural correlations induced by sparse channels and adaptively aggregates the informative features into a compact token. On the other hand, a simple MLP network is adopted to compress the beamformer label since it is a dense optimization outputs with no clear local structures.

\subsubsection{Pilot EDN} \label{pilot_edn_arch}

We first describe the pilot EDN architecture. Given the received pilot signal $\bm{Y} \in \mathbb{C}^{N \times L_p}$, we stack its real and imaginary parts to obtain $\bm{y} = \text{vec}([\Re(\bm{Y});\, \Im(\bm{Y})]) \in \mathbb{R}^{2NL_p}$. As shown in Fig.~\ref{icl_net_arch}, the pilot encoder $\mathcal{E}_p: \mathbb{R}^{2NL_p} \to \mathbb{R}^d$ maps $\bm{y}$ to a compact pilot token $\bm{z} \in \mathbb{R}^d$. The encoder network consists of two 1D convolutional modules, an attention-pooling module, and a two-layer MLP module, where each MLP layer includes a fully-connected (FC) component, a GELU activation component, and a LayerNorm component. 
To pretrain the pilot encoder, we employ another three-layer MLP network as an auxiliary channel decoder $\mathcal{D}_c: \mathbb{R}^{d} \to \mathbb{R}^{2KN}$, which reconstructs the real-valued channel vector $\bm{h}$ from the pilot token $\bm{z}$. Note that this auxiliary decoder is used only during pretraining to encourage the pilot encoder to extract channel-informative latent features from compressed pilot signals, instead of performing explicit CSI reconstructions. 

We then pretrain the pilot EDN using the multi-scenario labeled datasets in \eqref{ini_dataset_def}, which allow the encoder networks to learn shared token representations across different channel models. Denote the trainable parameters of the pilot encoder and the auxiliary channel decoder as $\bm{\phi}_p$ and $\bm{\psi}_c$, respectively. The pilot EDN is trained by minimizing the following channel reconstruction loss
\begin{align} \label{pilot_edn_loss}
\min_{\bm{\phi}_p,\bm{\psi}_c} \ \sum_{s \in \mathcal{S}} \mathbb{E}_{(\bm{H},\bm{Y})\in \mathcal{V}_s^{(0)}} \bigl\| \bm{h} - \mathcal{D}_c\bigl(\mathcal{E}_p(\bm{y})\bigr) \bigr\|_2^2,
\end{align}
where $\bm{h}=\text{vec}([\Re(\bm{H});\Im(\bm{H})]) \in \mathbb{R}^{2NK}$ and $\bm{y}=\text{vec}([\Re(\bm{Y});\Im(\bm{Y})]) \in \mathbb{R}^{2N L_p}$. After convergence, the trained channel decoder $\mathcal{D}^{*}_c$ is discarded while the trained pilot encoder $\mathcal{E}^{*}_p$ is frozen and used for pilot-token construction in the subsequent training stages.


\subsubsection{Beamformer EDN} \label{bf_edn_arch}

We now describe the beamformer EDN architecture. Given a beamformer matrix
$\boldsymbol W\in\mathbb C^{N\times K}$, we stack its real and imaginary parts
to obtain $\bm{w} = \text{vec}([\Re(\bm{W});\, \Im(\bm{W})]) \in \mathbb{R}^{2NL_p}$. As shown in Fig.~\ref{icl_net_arch}, the beamformer encoder
$\mathcal E_b:\mathbb R^{2NK}\rightarrow\mathbb R^d$ compresses $\boldsymbol w$ into a beamformer token, i.e., $\boldsymbol c=\mathcal E_b(\boldsymbol w)\in\mathbb R^d$. Both encoder $\mathcal E_b$ and decoder $\mathcal D_b$ are implemented as three-layer MLP networks. The beamformer decoder $\mathcal D_b:\mathbb R^d\rightarrow\mathbb R^{2NK}$ then maps a compressed beamformer token back to the real-valued beamformer vector. The reconstructed vector is first normalized to satisfy the transmit power constraint $\|\boldsymbol w\|^2=P$ and then reshaped into a complex beamformer matrix $\bm{W}$.


The pretraining of beamformer EDN is similar to that of pilot EDN. Denote the trainable parameters of the beamformer encoder and beamformer decoder as $\bm{\phi}_b$ and $\bm{\psi}_b$, respectively. The beamformer EDN is pretrained by minimizing the reconstruction loss
\begin{align} \label{bf_edn_loss}
\min_{\bm{\phi}_b,\bm{\psi}_b} \ \sum_{s\in\mathcal{S}} \mathbb{E}_{\bm{W}^{\text{m}} \in \mathcal{V}_s^{(0)}} \bigl\| \bm{w}^{\text{m}} - \mathcal{D}_b \bigl(\mathcal{E}_b(\bm{w}^{\text{m}})\bigr) \bigr\|^2,
\end{align}
where $\bm{w}^{\text{m}}=\text{vec}([\Re(\bm{W}^{\text{m}});\Im(\bm{W}^{\text{m}})])$. After convergence, both the trained beamformer encoder $\tilde{\mathcal{E}}_b$ and the trained beamformer decoder $\tilde{\mathcal{D}}_b$ are retained as the initializations for the subsequent training stages.



\section{Enhanced ICL for Multi-Model Pilot-based Beamforming} \label{evolve_icl_framework}

Building on the pretrained EDNs, we now present the enhanced ICL framework by first detailing the ICL Transformer architecture and then presenting our proposed curriculum self-evolving training scheme.




\subsection{ICL Transformer Network} \label{icl_tf_backbone}


For each pilot-beamformer sample $(\boldsymbol Y_j,\hat{\boldsymbol W}_j)$ in the context dataset, the pilot encoder (pretrained and frozen) and beamformer encoder (remain trainable) produce the compressed tokens as follows
\begin{align} \label{token_generate_edn}
\bm{z}_j = \mathcal{E}^{*}_p(\bm{Y}_j) \in \mathbb{R}^d, \quad
\hat{\bm{c}}_j = \mathcal{E}_b(\hat{\bm{W}}_j) \in \mathbb{R}^d.
\end{align}
Given a query input $\bm{Y}_{\ell+1}$, its pilot token is denoted by $\bm{z}_{\ell+1} = \mathcal{E}^{*}_p(\bm{Y}_{\ell+1})$. The ICL token sequence is then constructed as
\begin{align}\label{compress_icl_sequence}
\mathcal{Z}^{\ell} = [(\bm{z}_1, \hat{\bm{c}}_1), (\bm{z}_2, \hat{\bm{c}}_2), \ldots, (\bm{z}_\ell, \hat{\bm{c}}_\ell), \bm{z}_{\ell+1}] \in \mathbb R^{d\times L_\text{seq}}, 
\end{align}
where $\{(\bm{z}_j, \hat{\bm{c}}_j)\}_{j=1}^\ell$ are the context tokens and $\bm{z}_{\ell+1}$ is the query input token. The sequence length is $L_\text{seq} \triangleq 2\ell + 1$.


A Transformer network with $T$ layers is employed to process the token sequence $\mathcal{Z}^{\ell}$ in \eqref{compress_icl_sequence}. 
First, each token is projected to a higher-dimensional embedding space: 
\begin{align} \label{input_embed}
\bm{X}^{(0)} =  \bm{\mathcal{W}}_{\mathrm{in}} \mathcal{Z}^{\ell} + \bm{b}_{\mathrm{in}} \bm{1}_{L_\text{seq}}^T  \in \mathbb{R}^{\alpha \times L_\text{seq}},
\end{align}
where the dimension $\alpha > d$, $\bm{1}_{L_\text{seq}}$ is an all-one vector, $\bm{\mathcal{W}}_{\mathrm{in}} \in \mathbb{R}^{\alpha \times d}$ and $\bm{b}_{\mathrm{in}} \in \mathbb{R}^{\alpha}$ are learnable parameters. Starting from $\boldsymbol X^{(0)}$, the hidden states are updated for $t=1,\ldots,T$ as
\begin{align}
\bar{\bm{X}}^{(t-1)} &= \mathrm{LayerNorm} \bigl(\bm{X}^{(t-1)}\bigr), \label{prenorm_1}\\
{\bm{Z}}^{(t)} &= \bm{X}^{(t-1)} + \mathrm{MHSA}^{(t)}\!\bigl(\bar{\bm{X}}^{(t-1)}\bigr), \label{mhsa_res}\\
\bar{\bm{Z}}^{(t)} &= \mathrm{LayerNorm} \bigl({\bm{Z}}^{(t)}\bigr), \label{prenorm_2}\\
\bm{X}^{(t)} &= {\bm{Z}}^{(t)} + \mathrm{MLP}^{(t)}\!\bigl(\bar{\bm{Z}}^{(t)}\bigr). \label{ffn_res}
\end{align}
We next specify the multi-head self-attention module in \eqref{mhsa_res} at the $t^{\text{th}}$ layer. For the $e^{\text{th}}$ attention head, the query, key, and value matrices are computed as
\begin{align} \label{eq_qkv}
\bm{Q}_e^{(t)} &= (\bar{\bm{X}}^{(t-1)})^T \bm{R}_e^{Q,(t)}, \notag \\ 
\bm{K}_e^{(t)} &= (\bar{\bm{X}}^{(t-1)})^T \bm{R}_e^{K,(t)}, \notag \\ 
\bm{V}_e^{(t)} &= (\bar{\bm{X}}^{(t-1)})^T \bm{R}_e^{V,(t)},
\end{align}
where $\bm{R}_e^{Q,(t)}, \bm{R}_e^{K,(t)}, \bm{R}_e^{V,(t)} \in \mathbb{R}^{\alpha \times D_e}$ denote the learnable weight matrices, $D_e = D/E$ denotes the head dimension, and $E$ denotes the number of attention heads. The attention output of the $e^{\text{th}}$ head is computed as
\begin{align} \label{attn_score_compute}
\bm{A}_e^{(t)} = \mathrm{softmax} \left( \frac{\bm{Q}_e^{(t)} (\bm{K}_e^{(t)})^T}{\sqrt{D_e}} \right)\bm{V}_e^{(t)} \ \in \ \mathbb{R}^{L_\text{seq} \times D_e}.
\end{align}
The multi-head attention output is obtained by concatenating all heads and projecting the result back to the embedding dimension:
\begin{align} \label{mhsa_final_output}
\mathrm{MHSA}(\bar{\bm{X}}^{(t-1)}) = \bm{R}_{\text{proj}}^{(t)} \ [\bm{A}_1^{(t)}, \ldots, \bm{A}_E^{(t)}]^T,
\end{align}
where $\bm{R}_{\text{proj}}^{(t)} \in \mathbb{R}^{\alpha \times D}$ is the learnable output projection matrix.


After the final Transformer layer, the hidden state is vectorized and mapped to the predicted query beamformer token:
\begin{align}\label{output_proj_bf_sol}
\bm{x}^{(T)} &= \text{vec}(\bm{X}^{(T)}) \in \mathbb{R}^{\alpha L_\text{seq}}, \notag \\ 
\bm{c} &= \bm{\mathcal{W}}_{\mathrm{o}} \cdot \bm{x}^{(T)} + \bm{b}_{\mathrm{o}} \in \mathbb{R}^d,
\end{align}
where $\bm{\mathcal{W}}_{\mathrm{o}} \in \mathbb{R}^{d \times \alpha L_\text{seq}}$ and $\bm{b}_{\text{o}} \in \mathbb{R}^{d}$ are learnable parameters. The predicted token $\bm{c}$ is then decoded into a real-valued beamformer vector:
\begin{align}\label{final_bf_decode}
\bm{w} = \mathcal{D}_b(\bm{c}) \in \mathbb{R}^{2NK}.
\end{align}
Finally, $\bm{w}$ is first normalized by $\bm{w}=\frac{\sqrt{P} \cdot \bm{w}}{\|\bm{w}\|}$, and then reshaped into a complex beamformer matrix:
\begin{align} \label{bf_sol_reshape}
\bm{W} &= \text{vec}^{-1}(\bm{w}[1:NK]) + j \cdot \text{vec}^{-1}(\bm{w}[NK+1:2NK]).
\end{align}


Importantly, the Transformer blocks do not share parameters across
layers. We first collect the learnable parameters in the $t^{\text{th}}$ Transformer
layer as
\begin{equation}
\begin{aligned}
\bm{\theta}_t
=
\Bigl\{
&
\{\bm R_{e}^{Q,(t)},\bm R_{e}^{K,(t)},\bm R_{e}^{V,(t)}\}_{e=1}^{E},
\bm R_{\text{proj}}^{(t)}, \\
&
\bm\beta_{\rm MLP}^{(t)},
\bm\beta_{{\rm LN},1}^{(t)},
\bm\beta_{{\rm LN},2}^{(t)}
\Bigr\},
\end{aligned}
\end{equation}
where $\bm\beta_{\rm MLP}^{(t)}$ denotes the parameters of the MLP network in \eqref{ffn_res},
$\bm\beta_{{\rm LN},1}^{(t)}$ and
$\bm\beta_{{\rm LN},2}^{(t)}$ denote the learnable affine parameters of the
two LayerNorm modules in \eqref{prenorm_1} and \eqref{prenorm_2}. The complete set of
learnable parameters in the ICL Transformer is then given by
\begin{equation}
\begin{aligned} \label{theta_summary}
\bm{\theta}
=
\Bigl\{
&
\bm{\mathcal W}_{\rm in},\bm b_{\rm in},
\bm{\mathcal W}_{\rm o},\bm b_{\rm o}
\Bigr\}
\cup
\bigcup_{t=1}^{T}\bm{\theta}_t.
\end{aligned}
\end{equation}

For the beamformer encoder and decoder networks, the learnable parameters are given as follows
\begin{align} \label{edn_para_sumary}
\bm{\phi}_b = \bigcup_{i=1}^{3} \Big\{ \bm{\xi}^{(i)}_{\text{FC}} \Big\}, \ \bm{\psi}_b = \bigcup_{i=1}^{3} \Big\{ \bm{\zeta}^{(i)}_{\text{FC}} \Big\},
\end{align}
where $\bm{\xi}^{(i)}_{\text{FC}}$ denotes the parameters of the $i^{\text{th}}$ linear layer of the beamformer encoder, and $\bm{\zeta}^{(i)}_{\text{FC}}$ denotes the parameters of the $i^{\text{th}}$ linear layer of the beamformer decoder.


\subsection{Curriculum Learning with Context Bootstrapping} \label{self_evolve_curriculum}

As discussed in Sec.~\ref{prob_state}, curriculum learning (CL) plays a central role in the proposed enhanced ICL framework. It addresses two challenges in ICL beamforming: highly non-convex optimization and efficient construction of informative context datasets. First, CL provides a smooth transition from supervised imitation to unsupervised sum-rate maximization. This transition anchors the model in a well-initialized region in the early stage and mitigates poor convergence to local optima in the sum-rate maximization. Second, the self-evolving context dataset reduces the need for a large precomputed labeled dataset. Instead of relying on fixed beamformer labels, the context dataset is adaptively expanded and refined during training. As high-quality model-generated samples are gradually incorporated, the context distribution becomes broader and more informative, enabling the ICL model to condition on more diverse demonstrations and generalize better across the problem family.


Based on the architecture in Fig.~\ref{icl_net_arch}, the complete training procedure consists of three stages: EDN pretraining, supervised warm-start, and CL with evolving context. The EDN pretraining has been introduced in Sec.~\ref{edn_pretrain_sec}. In the following, with the pretrained pilot encoder fixed, we focus on the joint training of the ICL Transformer and the beamformer EDN through the latter two stages.

\subsubsection{Supervised Warm-Start} \label{warm_start_phase}

The \emph{supervised warm-start} training stage jointly trains the ICL Transformer, the beamformer encoder and decoder, by using the LMMSE-labeled context dataset $\mathcal V^{(0)}$ in \eqref{ini_dataset_def}. This stage provides a stable initialization before introducing the non-convex sum-rate objective. 

Each training batch contains the same number of samples from all channel models $s \in \mathcal{S}$. For each training sample from model $s$, we randomly select $\ell$ pilot-beamformer pairs from $\mathcal V_s^{(0)}$ to form the context $\mathcal{C}_s=[(\bm{Y}_{j},\bm{W}^{\text{m}}_{j})]_{j=1}^{\ell}$, and select another $(\bm{Y}_{\ell+1},\bm{W}^{\text{m}}_{\ell+1}) \in \mathcal{V}_s^{(0)}$ as the query pair \footnote{The query pair should be excluded from the context pairs for all training samples to prevent data-leakage.}. 
Following \eqref{token_generate_edn} and \eqref{compress_icl_sequence}, the corresponding compressed token sequence is constructed as $\mathcal{Z}_s^{\ell}$. The token sequence is then processed by the ICL Transformer, and the predicted beamformer token is decoded by the beamformer decoder:
\begin{align} \label{forward_tf_dec}
\bm{c} = \mathcal{T}_b(\mathcal{Z}_s^{\ell}), \ \bm{w} = \mathcal{D}_b(\bm{c}),
\end{align}
where $\mathcal{T}_b$ denotes the ICL Transformer with learnable parameters $\bm{\theta}$ in \eqref{theta_summary}, and the corresponding reshaped and power-normalized beamformer matrix is denoted by $\boldsymbol W(\bm{Y}_{\ell+1},\mathcal{C}_s)$. After randomly initializing $\mathcal{T}_b$ and initializing $(\mathcal{E}_b, \mathcal{D}_b)$ by the pretrained models $(\tilde{\mathcal{E}}_b, \tilde{\mathcal{D}}_b)$ obtained in Sec.~\ref{bf_edn_arch}, the following supervised training across all channel models is performed:
\begin{align} \label{warm_start_loss}
\min_{\bm{\theta},\bm{\phi}_b,\bm{\psi}_b} \  \sum_{s \in \mathcal{S}}\mathbb{E}_{\{(\bm{Y}_{j},\bm{W}^{\text{m}}_{j})\}_{j=1}^{\ell+1} \in \mathcal{V}_s^{(0)}} \bigl\| \boldsymbol W(\bm{Y}_{\ell+1},\mathcal{C}_s) - \bm{W}^{\text{m}}_{\ell+1} \bigr\|_F^2.
\end{align}
By imitating the LMMSE beamformers, a reliable initialization for the subsequent CL-based training is provided. 


\subsubsection{CL with Context Bootstrapping} \label{self_evolve_cl}


After the supervised warm-start, the learning model gradually transitions to the sum-rate maximization. 
Let $\mathcal V_s$ denote the evolving context dataset under channel model $s$, initialized as $\mathcal V_s^{(0)}$ and bootstrapped during training. Now consider each element $(\boldsymbol H,\boldsymbol Y,\hat{\boldsymbol W}) \in \mathcal V_s$. Initially, $\hat{\boldsymbol W}$ is the LMMSE solution $\bm{W}^{\text{m}}(\bm{H})$ in $\mathcal V_s^{(0)}$; later in the dataset $\mathcal V_s$ during bootstrapping, it may also be an accepted model-generated beamformer solution.

For notational clarity, define the model output under a query input $\bm{Y}$ and a context $\mathcal C_s \in \mathcal V_s$ as follows
\begin{align} \label{out_bf_sol_tf}
\boldsymbol W(\boldsymbol Y,\mathcal C_s) = \mathcal D_b \left( \mathcal T_b \left( \mathcal Z_s^\ell(\mathcal C_s,\boldsymbol Y) \right) \right).
\end{align}
We define the supervised imitation loss as follows:
\begin{align} \label{mse_loss_def}
L^{\text{sup}}(\bm{Y}, \hat{\bm{W}}, \mathcal C_s) = \bigl\| \boldsymbol W(\boldsymbol Y,\mathcal C_s) - \hat{\bm{W}} \bigr\|_F^2,
\end{align}
and define the sum-rate loss as:
\begin{align} \label{rate_loss_def}
L^{\text{rate}}(\bm{H},\bm{Y},\mathcal C_s) = - R_{\text{sum}}(\bm{H},\boldsymbol W(\boldsymbol Y,\mathcal C_s)).
\end{align}

At a certain epoch, let $r_u \in [0,1]$ denote the proportion of unsupervised samples within one training batch, which \emph{increases in a stepwise manner} and ends up with $r_u=1$ during CL-based training. 
The training objective under channel model $s$ can then be written in the following expected loss form:
\begin{align}  \label{hybrid_loss_obj}
\min_{\bm{\theta},\bm{\phi}_b,\bm{\psi}_b} \ \Big[ &(1-r_u)\, \mathbb E_{\substack{ (\boldsymbol H,\boldsymbol Y,\hat{\boldsymbol W})\in \mathcal V_s\\ \mathcal C_s \in \mathcal V_s}} \left[ L^{\text{sup}} (\boldsymbol Y,\hat{\boldsymbol W},\mathcal C_s) \right] \nonumber\\ &+ \gamma \cdot r_u \Big( \eta_d\, \mathbb E_{\substack{ (\boldsymbol H,\boldsymbol Y,\hat{\boldsymbol W})\in \mathcal V_s\\ \mathcal C_s \in \mathcal V_s}} \left[ L^{\text{rate}} (\boldsymbol H,\boldsymbol Y,\mathcal C_s) \right] \nonumber\\ & +(1-\eta_d)\, \mathbb E_{\substack{ \boldsymbol H \sim p_s, \bm{N}\\ \boldsymbol Y=\boldsymbol H \boldsymbol\Psi+\boldsymbol N\\ \mathcal C_s \in \mathcal V_s}} \left[ L^{\text{rate}} (\boldsymbol H,\boldsymbol Y,\mathcal C_s) \right] \Big) \Big], 
\end{align}
where $\gamma$ is a scaling factor that balances the supervised and sum-rate losses. To strike a balance between exploitation, i.e., refining existing samples in the dataset, and exploration, i.e., improving on newly generated samples, the unsupervised samples are further divided into those drawn from $\mathcal V_s$, which corresponds to the second term in \eqref{hybrid_loss_obj}, and those generated from the simulator, which corresponds to the third term in \eqref{hybrid_loss_obj}. $\eta_d\in[0,1]$ controls the ratio between these two components.

During CL-based training, the model continuously generates candidate beamformer solutions for newly sampled pilot signals. Specifically, under channel model $s$, we consider a fresh sample $(\tilde{\bm{H}}, \tilde{\bm{Y}}, \tilde{\bm{W}}^\text{m})$ generated by the simulator, where $\tilde{\bm{W}}^\text{m}$ is the LMMSE beamformer of $\tilde{\bm{H}}$, together with a randomly selected context $\mathcal C_s$ from $\mathcal V_s$. The model produces $\tilde{\bm{W}}(\tilde{\bm{Y}},\mathcal{C}_s)$ based on \eqref{out_bf_sol_tf},
with achieved sum rate
\begin{align} \label{new_sample_rate}
\tilde{R} = R_{\mathrm{sum}}(\tilde{\bm{H}}, \tilde{\bm{W}}(\tilde{\bm{Y}},\mathcal{C}_s)).
\end{align}
The candidate sample is admitted into the evolving context dataset only if it outperforms a prescribed fraction of the LMMSE performance on the same channel:
\begin{align} \label{per_instance_condi}
\tilde{R} \geq \alpha \cdot R_{\mathrm{sum}}(\tilde{\bm{H}}, \tilde{\bm{W}}^\text{m}),
\end{align}
where $\alpha$ increases from $\alpha_s$ to $\alpha_e$ during training. This instance-wise admission rule prevents low-quality model-generated solutions from polluting the context dataset. The threshold is intentionally relaxed in the early stage to encourage context diversity and becomes progressively stricter to improve the quality of retained demonstrations. Once accepted, the sample $(\tilde{\boldsymbol H},\tilde{\boldsymbol Y}, \tilde{\boldsymbol W}(\tilde{\bm{Y}},\mathcal{C}_s))$ is added to $\mathcal V_s$, and the generated beamformer $\tilde{\boldsymbol W}(\tilde{\bm{Y}},\mathcal{C}_s)$ is used in the later context constructions.

In addition to \textbf{adding qualified new samples}, each context dataset also keeps \textbf{revisiting and refining} its existing samples to improve the context quality. Specifically, given a sampled triplet $(\boldsymbol H,\boldsymbol Y,\hat{\boldsymbol W})\in \mathcal V_s$, if the model generates a beamformer solution $\bm{W}(\bm{Y},\mathcal{C}_s)$ during training, such that 
\begin{align} \label{revisit_update_condi}
R_{\mathrm{sum}}(\bm{H},\bm{W}(\bm{Y},\mathcal{C}_s)) > R_{\mathrm{sum}}(\bm{H},\hat{\boldsymbol W}),
\end{align}
$(\boldsymbol H,\boldsymbol Y,\hat{\boldsymbol W})$ will be replaced by $(\boldsymbol H,\boldsymbol Y,\bm{W}(\bm{Y},\mathcal{C}_s))$ in $\mathcal V_s$. 

To conclude, unlike prior ICL beamforming scheme~\cite{wen2026icwlm} that requires a large-scale WMMSE-labeled dataset for context construction, the enhanced ICL scheme only requires precomputing a small seed dataset with inexpensive LMMSE labels, which is enabled by the context bootstrapping mechanism during CL-based training.

\begin{algorithm}[t]
\caption{Enhanced ICL Pilot-based Beamforming Scheme under Multiple Channel Models.}
\label{alg_1}
\begin{algorithmic}[1]
\small
\setlength{\lineskip}{2pt}
\STATE \textbf{Parameters}: epoch number of EDN pretraining stage $E_0$, epoch number of warm-start stage $E_1$, epoch number of CL stage $E_2$, sample ratios $(r_u,\eta_d)$, scaling factor $\gamma$, context length $\ell$
\STATE \textbf{Training:}
\STATE \textbf{EDN pretraining:}
\STATE \quad Build $\mathcal{V}^{(0)} = \bigcup_{s \in \mathcal{S}} \mathcal{V}_s^{(0)}$ with LMMSE labels
\STATE \quad \textbf{for} epoch $e = 1, \ldots, E_0$ \textbf{do}
\STATE \quad \quad Construct a training batch containing the samples from \\
\quad \quad all channel models
\STATE \quad \quad Train $(\mathcal{E}_p, \mathcal{D}_c)$ on $\mathcal{V}^{(0)}$ via~\eqref{pilot_edn_loss} 
\STATE \quad \quad Train $(\mathcal{E}_b, \mathcal{D}_b)$ on $\mathcal{V}^{(0)}$ via~\eqref{bf_edn_loss}
\STATE \quad \textbf{end for}
\STATE \quad Discard $\mathcal{D}^{*}_c$, freeze $\mathcal{E}^{*}_p$, retain $\tilde{\mathcal{E}}_b$ and $\tilde{\mathcal{D}}_b$
\STATE \textbf{Supervised warm-start:}
\STATE \quad \textbf{for} epoch $e = 1, \ldots, E_1$ \textbf{do}
\STATE \quad \quad Construct a training batch containing the samples from \\
\quad \quad all channel models
\STATE \quad \quad Draw $\ell$ context pairs and a query pair from $\mathcal{V}^{(0)}_s$, \\
\quad \quad compress them via $\mathcal{E}^{*}_p$ and $\mathcal{E}_b$
\STATE \quad \quad Construct the input token sequence $\mathcal{Z}_s^{\ell}$
\STATE \quad \quad Perform the supervised training based on \eqref{warm_start_loss}
\STATE \quad \textbf{end for}
\STATE \textbf{CL with context bootstrapping:}
\STATE \quad Initialize $\mathcal{V}_s=\mathcal{V}^{(0)}_s$ for all models $s \in \mathcal{S}$
\STATE \quad \textbf{for} epoch $e = E_1+1, \ldots, E_1+E_2$ \textbf{do}
\STATE \quad \quad Update $r_u$ according to the curriculum
\STATE \quad \quad Construct a training batch containing the samples from \\
\quad \quad all channel models
\STATE \quad \quad Based on $(r_u,\eta_d)$, select samples from $\mathcal{V}_s$ and \\
\quad \quad generate fresh samples from the simulator
\STATE \quad \quad Perform the CL-based training using \eqref{hybrid_loss_obj}
\STATE \quad \quad Add each fresh sample into $\mathcal{V}_s$ if it satisfies \eqref{per_instance_condi}
\STATE \quad \quad Update each existing sample in $\mathcal{V}_s$ if it satisfies \eqref{revisit_update_condi}
\STATE \quad \textbf{end for}
\STATE Obtain the trained ICL model $(\mathcal{E}^*_p,\mathcal{E}^*_b,\mathcal{T}^*_b,\mathcal{D}^*_b)$ and the \\ 
bootstrapped datasets $\{\mathcal{V}_s\}_{s \in \mathcal{S}}$ for all channel models
\vspace{4pt}
\hrule
\vspace{4pt}
\STATE \textbf{Inference:}
\STATE \quad Receive the new pilot signal $\bm{Y}$
\STATE \quad Given channel model $s$, randomly sample $\mathcal{C}_s$ from $\mathcal{V}_s$ 
\STATE \quad Construct token sequence $\mathcal{Z}_s^\ell$ and generate the solution
\end{algorithmic}
\end{algorithm}

\section{Extension to Model-mismatch Settings} \label{exten_model_mismatch}

The previous sections consider an aligned setting, where the pilot signal and the beamforming objective are associated with the same channel realization. We now extend the proposed ICL framework to more practical model-mismatch settings, in which the received pilot signal is generated from a channel state or propagation condition that is not fully aligned with the target channel used for beamformer evaluation. We first introduce several representative sources of mismatch, and then develop a mismatch-aware ICL scheme that incorporates these mismatches into the general framework.

\subsection{Model Mismatches} \label{mm_types}

\subsubsection{Uplink-Downlink Mismatch} \label{ul_dl_mismatch}

In FDD systems, the uplink and downlink channels share similar topology-related parameters, but operate at different carrier frequencies~\cite{Zhong2020FDD}. Therefore, frequency-dependent parameters may differ between the uplink and downlink. Accordingly, we model them as two correlated sparse geometric channels:
\begin{align} \label{ul_dl_chan_model}
\bm h_k^{\mathrm{ul}}&
=\sqrt{\frac{N}{L}}\sum_{p=1}^{L}\alpha_{k,p}^{\mathrm{ul}}\bm a_{f_{\mathrm{ul}}}(\phi_{k,p}),
\notag \\
\bm h_k^{\mathrm{dl}}&=
\sqrt{\frac{N}{L}}
\sum_{p=1}^{L}
\alpha_{k,p}^{\mathrm{dl}}
\bm a_{f_{\mathrm{dl}}}(\phi_{k,p}),
\end{align}
where $f_{\mathrm{ul}}$ and $f_{\mathrm{dl}}$ denote the uplink and downlink carrier frequencies, respectively, and the steering vector is given by
\begin{align}
\bm a_f(\phi)=\frac{1}{\sqrt{N}} \left[
1,
e^{\text{j} 2\pi f d_0 \sin\phi/c},
\ldots,
e^{\text{j} 2\pi f d_0 (N-1)\sin\phi/c}
\right]^T,
\end{align}
where $f$ is the carrier frequency, $c$ is the speed of light, and $d_0$ is the antenna spacing. The path gains in \eqref{ul_dl_chan_model} are typically modeled as the following correlated perturbation model~\cite{Zhong2020PartialReciprocity}:
\begin{align} \label{gain_ul_dl}
\alpha_{k,\ell}^{\mathrm{dl}}
=
\rho_{\mathrm{ud}}\alpha_{k,\ell}^{\mathrm{ul}}
+
\sqrt{1-\rho_{\mathrm{ud}}^2}\epsilon_{k,\ell}^{\mathrm{ud}},
\end{align}
where $\rho_{\mathrm{ud}}\in[0,1]$ controls the uplink-downlink correlation, and $\epsilon_{k,\ell}^{\mathrm{ud}}\sim\mathcal{CN}(0,1)$ is an independent perturbation. In this setting, the pilot signal $\bar{\bm{Y}}$ is computed by \eqref{pilot_sig_model} based on the uplink channel $\bm{H}^{\text{ul}}=[\bm{h}_1^{\text{ul}},\dots,\bm{h}_K^{\text{ul}}]$.



\subsubsection{Channel-Aging Mismatch} \label{time_csi_mismatch}

We next consider the channel-aging mismatch. If the processing delay from pilot reception to beamformer generation exceeds the channel coherence time, the received pilot signal is generated from an outdated channel, whereas the beamformer is applied to an updated channel. Specifically, let $\bar{\bm{h}}_k$ and $\bm{h}_k$ denote the outdated and updated channels of user $k$, respectively. Since the path topology and angular support usually vary slowly over a short delay, we keep the same AoDs and model the temporal variation through Doppler-induced path-gain evolution:
\begin{align} \label{time_mismatch_channel}
\bar{\bm{h}}_k
&=
\sqrt{\frac{N}{L}}
\sum_{\ell=1}^{L}
\alpha_{k,\ell} \bm a(\phi_{k,\ell}), \notag \\
\bm{h}_k
&=
\sqrt{\frac{N}{L}}
\sum_{\ell=1}^{L}
\left( \rho
\alpha_{k,\ell} e^{ \text{j} 2\pi \nu_{k,\ell}\Delta}
+ \sqrt{1-\rho^2}
\xi_{k,\ell}
\right)
\bm a(\phi_{k,\ell}),
\end{align}
where $\alpha_{k,\ell} \sim \mathcal{CN}(0,1)$, $\nu_{k,\ell}$ is the Doppler shift, $\Delta$ is the processing delay of the BS, and $\xi_{k,\ell} \sim \mathcal{CN}(0,1)$ captures residual fading uncertainty. 
In this setting, the pilot signal $\bar{\bm{Y}}$ is computed by \eqref{pilot_sig_model} based on the outdated CSI $\bar{\bm{H}}=[\bar{\bm{h}}_1,\dots,\bar{\bm{h}}_K]$.

\subsubsection{Pilot-Contamination Mismatch} \label{pilot_contami}

We further consider the mismatch induced by pilot contamination~\cite{Marzetta2010}. When pilot resources are limited, users in neighboring cells may adopt non-orthogonal pilot sequences. As a result, the pilot signal received at the target BS is contaminated by inter-cell interference. 
Specifically, let $\bm H^{\text{t}}$ denote the target-cell channel, and let $\bm H_i^{\rm c}$ denote the contaminating channel from the $i^{\text{th}}$ neighboring cell, where $i\in\mathcal B_{\rm c}$ and $\mathcal B_{\rm c}$ is the set of interfering cells. Then the pilot signal can be modeled as
\begin{align} \label{pc_observation}
\bar{\bm{Y}}=\bm H^{\text{t}}\bm\Psi+\sum_{i\in\mathcal B_{\rm c}} \sqrt{\eta_i}\bm H^{\rm c}_i\bm\Phi_i+\bm N,
\end{align}
where $\bm\Psi$ is the pilot matrix used by the target users, $\bm\Phi_i$ is the pilot matrix used by the $i^{\text{th}}$ cell, and $\eta_i$ controls the relative contamination strength. 


\subsection{Mismatch-Aware ICL Scheme} \label{icl_for_mismatch}

We now propose a mismatch-aware ICL scheme to handle the mismatch cases introduced in Sec.~\ref{mm_types}. Consider a set of channel models $\mathcal{S}$ without mismatches. For each $s \in \mathcal{S}$, let $s^{(\text{UD})}$, $s^{(\text{CA})}$, and $s^{(\text{PC})}$ denote the observed models $s$ induced by uplink-downlink mismatch, channel-aging mismatch, and pilot contamination mismatch, respectively. Taking the uplink-downlink mismatch as an example,
the initial context dataset is constructed as
\begin{align} \label{unified_mismatch_dataset}
\mathcal V_{s,\text{UD}}^{(0)} = \left\{(\bm{H}_j,\bar{\bm{Y}}_j,\bm W_j^{\text{m}}) \right\}_{j=1}^{M_0},
\end{align}
where $\bm{H}_j \sim p_s$ is the target channel, $\bar{\bm{Y}}_j$ is the pilot signal generated under the mismatched model $s^{(\text{UD})}$, and $\bm W_j^{\text{m}}$ is the LMMSE beamformer computed from $\bm{H}_j$. 

Let $\mathcal{I} = \{\text{UD},\text{CA},\text{PC}\}$ denote the set of mismatch types. For each $s\in\mathcal{S}$ and each $i\in\mathcal I$, denote by $q_{s,i}$ the distribution of the mismatched pilot signal $\bar{\bm{Y}}$. Then the ultimate objective is to maximize the expected sum rate over all channel models $s \in \mathcal{S}$ and all mismatch types $i \in \mathcal{I}$:
\begin{align} \label{mismatch_icl_formu}
\max_{\bm{\theta}} \ \sum_{\substack{s \in \mathcal{S}, \\ i \in \mathcal{I}}} \
\mathbb E_{\substack{ \bm{H} \sim p_s, \ \bar{\bm{Y}} \sim q_{s,i} \\ \mathcal C_{s,i} \in \mathcal{V}_{s,i}(\bm{\theta})}}
\left[R_\text{sum} \left(\bm{H}, \mathcal{M}_{\bm{\theta}}(\mathcal{Z}_{s,i}^\ell)\right)\right],
\end{align}
where $\mathcal C_{s,i}=[(\bar{\bm{Y}}_j,\hat{\bm{W}}_j)]_{j=1}^\ell$ is sampled from the current context dataset $\mathcal{V}_{s,i}$, $\mathcal{Z}_{s,i}^\ell=[\mathcal C_{s,i},\bar{\bm{Y}}]$ is the input sequence, and $\mathcal{M}_{\bm{\theta}}$ is the learnable neural network. The other training strategies are similar to those described in Sec.~\ref{evolve_icl_framework}.

Equivalently, the no-mismatch ICL scheme learns to adapt to the channel models in $\mathcal{S}$, whereas the mismatch-aware ICL scheme extends the adaptation space to $\mathcal{S} \times \mathcal{I}$. By embedding the observation-target mismatch into the construction of demonstration pairs, the proposed scheme avoids explicit intermediate channel calibration and enables end-to-end mismatch-aware beamforming.

\section{Performance Evaluation} \label{perf_eval}

\subsection{Experiment Setup} \label{setup_sec}

\subsubsection{Network Parameters} \label{net_parameter}

We consider a MU-MISO downlink system where an $N$-antenna BS serves $K$ single-antenna users under the total transmit power $P$ and common noise variance $\sigma^2_1=\dots=\sigma^2_K=\sigma^2$. Unless otherwise specified, we set $K=N=32$, $P=1$, $\text{SNR}=\frac{1}{\sigma^2}$, the pilot length $L_p=20$, the number of demonstration pairs $\ell=5$, and input sequence length $L_{\text{seq}}=2\ell+1=11$.

The parameters of Algorithm~\ref{alg_1} are set as follows. The three training stages, namely EDN pretraining, supervised warm-start, and CL with context bootstrapping, are run for $E_0=100$, $E_1=50$, and $E_2=700$ epochs, respectively. The initial LMMSE-labeled context dataset contains $M_0=1024$ samples per channel model, and the final bootstrapped context dataset contains $M=5 \times 10^4$ samples per channel model. In \eqref{hybrid_loss_obj}, the scaling factor is set to $\gamma=0.1$, the unsupervised ratio $r_u$ is smoothly increased from $0$ to $1$ with step size $0.1$, and the ratio $\eta_d$ is set to $0.3$. The training batch size is $B_t=256$, and the test set size is $B_i=800$, containing samples from all $|\mathcal{S}|=8$ channel models. During training, the channel SNR is sampled from the set $\mathcal{K}_t=[5,10,15,20] \ \text{dB}$. The learning rate is initialized as $\eta = 10^{-3}$ and decayed to $\eta=10^{-4}$ following a cosine decay schedule. The admission thresholds in \eqref{per_instance_condi} are set to $\alpha_s=0.3$ and $\alpha_e=0.7$ \footnote{Note that the RHS in \eqref{per_instance_condi} is computed based on perfect CSI, but the LHS assumes that CSI is unavailable. Hence it is reasonable to set $\alpha_s < \alpha_e < 1$.}.

Moreover, Table.~\ref{tab_tf_net} summarizes the parameter sizes of the main components in the ICL network shown in Fig.~\ref{icl_net_arch}. The pilot encoder maps the pilot signal from dimension $2NL_p=1280$ to $d=64$, using convolutional kernel size $3$ and hidden dimension $D_h=256$. The beamformer encoder maps the beamformer from dimension $2NK=2048$ to $d=64$ with hidden dimension $D_h=256$, while beamformer decoder maps the beamformer from dimension $d=64$ to $2NK=2048$ with hidden dimension $D_h=256$. The ICL Transformer first embeds each token from $d=64$ to $\alpha=128$, processes the sequence through $T=6$ Transformer blocks, and finally projects the vectorized hidden state from dimension $\alpha L_{\rm seq}=1408$ to $d=64$. Each Transformer block contains two LayerNorm modules, an MHSA module with $E=8$ attention heads and head dimension $D_e=64$, and an MLP module with layer sizes $128 \times 256$, $256 \times 256$, and $256 \times 128$. Overall, the ICL-based neural network contains approximately $4.03 \times 10^6$ parameters. 




\begin{table}[t]
    \centering
    \caption{Parameter sizes of the main components in the ICL network.}
    \label{tab_tf_net}
    \begin{tabular}{c c c}
        \hline
        \textbf{Name} & \textbf{Notations} & \textbf{Size} \\
        \hline
        Pilot encoder & \(\mathcal E_p\) & \(344576\) \\
        Beamformer encoder & \(\mathcal E_b\) & \(606720\) \\
        Beamformer decoder & \(\mathcal D_b\) & \(606720\) \\
        Input embedding & \(\bm W_{\rm in},\bm b_{\rm in}\) & \(8320\) \\
        Weight matrices per block & 
        \(\{\bm R_e^{Q},\bm R_e^{K},\bm R_e^{V}\}_{e=1}^{E},\bm R_{\rm proj}\) 
        & \(263808\) \\
        MLP per block & \(\beta_{\rm MLP}\) & \(131712\) \\
        LayerNorm per block & \(\beta_{\rm LN,1},\beta_{\rm LN,2}\) & \(512\) \\
        Output projection & \(\bm W_o,\bm b_o\) & \(90176\) \\
        \hline
    \end{tabular}
\end{table}


\subsubsection{Multiple Sparse Channel Models}

We consider multiple cluster-based sparse channel models. Each channel model is specified by the triplet \((L_c,L_r,\sigma_\phi)\), where \(L_c\) denotes the number of scattering clusters, \(L_r\) denotes the number of rays per cluster, and \(\sigma_\phi\) denotes the intra-cluster angular spread. For each user $k$, the channel is generated as
$$
\bm h_k
=
\sqrt{\frac{N}{L_c L_r}}
\sum_{c=1}^{L_c}
\sum_{r=1}^{L_r}
\alpha_{k,c,r}\bm a(\phi_{k,c,r}),
$$
where \(\alpha_{k,c,r}\sim\mathcal{CN}(0,1)\). For a half-wavelength-spaced ULA, the steering vector is
$$
\bm a(\phi)
=
\frac{1}{\sqrt{N}}
\left[
1,e^{j\pi\sin\phi},\ldots,e^{j\pi(N-1)\sin\phi}
\right]^T .
$$
The ray angle is generated by
\begin{align}
\phi_{k,c,r}=\bar{\phi}_{k,c}+\Delta\phi_{k,c,r}, \ k\in[K], \ c \in [L_c], \ r \in [L_r],
\end{align}
where $\bar{\phi}_{k,c}\sim\text{Uniform}(-\pi/2,\pi/2)$ and $\Delta\phi_{k,c,r}\sim\mathcal{N}(0,\sigma_\phi^2)$. Therefore, different channel models are represented by different triplets \((L_c,L_r,\sigma_\phi)\), and the corresponding configurations are listed in Table~\ref{multi_scenarios}. 

\begin{table}[t]
\centering
\caption{Multiple Channel Model Configurations}
\label{multi_scenarios}
\small
\begin{tabular}{clccc}
\toprule
ID & Name & Clusters ($L_c$) & Rays ($L_r$) & Spread ($\sigma_\phi$)\\
\midrule
S1 & Dense Urban     & 3 & 5  & $10^\circ$ \\
S2 & LoS-Dominant    & 1 & 10 & $3^\circ$ \\
S3 & Rich Scatter    & 6 & 3  & $5^\circ$ \\
S4 & Suburban        & 2 & 8  & $15^\circ$ \\
S5 & Indoor Office   & 5 & 2  & $20^\circ$ \\
S6 & Near-LoS        & 1 & 15 & $2^\circ$ \\
S7 & Moderate Urban  & 4 & 4  & $8^\circ$ \\
S8 & Rayleigh        & / & /  & / \\
\bottomrule
\end{tabular}
\end{table}



\subsection{Ablation Study} \label{ablation_study}


We now present ablation studies on several key designs of the proposed enhanced ICL-based beamforming scheme, including the in-context architecture, curriculum learning (CL), and the context dataset bootstrapping. The following two non-learning baselines are used throughout simulations: 
\begin{itemize}
    \item \textbf{WMMSE-LS}: the BS first obtains the LS channel estimate from the received pilot signal, and then applies the iterative WMMSE algorithm in~\cite{shi2011wmmse} based on the estimated channel.
    \item \textbf{LMMSE-LS}: the BS first obtains the LS channel estimate from the received pilot signal, and then computes the LMMSE beamformer by~\eqref{mmse_bf} based on the estimated channel.
\end{itemize}
For ease of the result presentation, all ICL-based schemes in this subsection are trained over the multiple channel models listed in Table~\ref{multi_scenarios}, while the testing results are reported only on the selected ``Dense Urban'' channel model.

\begin{figure} [htbp]
    \centering
    \includegraphics[width=1.0\linewidth]{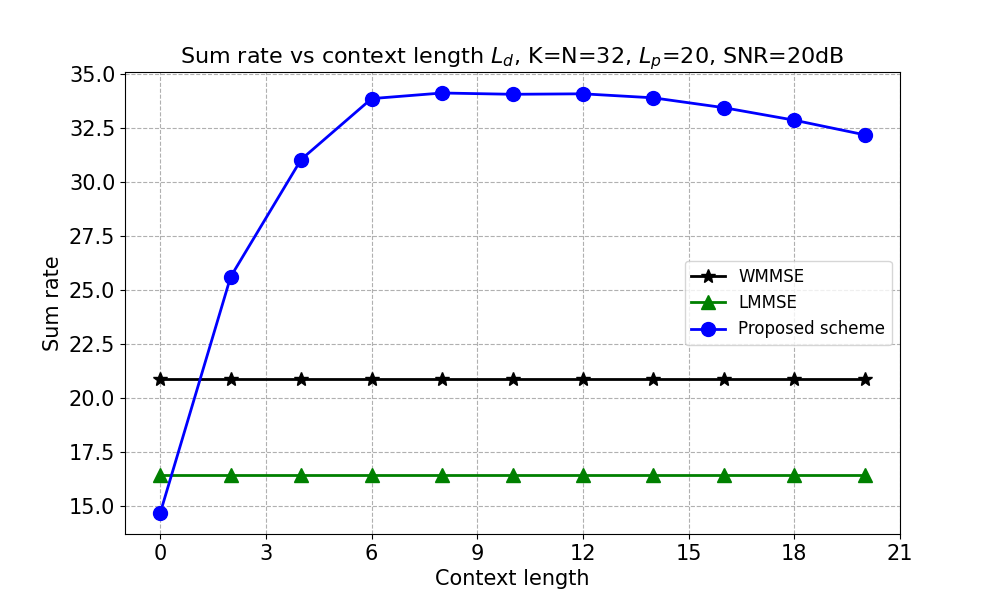}
    \caption{Performance under different context lengths $\ell$.}
    \label{context_ablation}
\end{figure}

We first evaluate the effect of the \textbf{context length} $\ell$ on the proposed ICL beamforming scheme. 
Fig.~\ref{context_ablation} plots the average testing sum rate versus $\ell$. It can be observed that the proposed scheme significantly outperforms both WMMSE-LS and LMMSE-LS baselines once demonstration context is provided. As $\ell$ increases from $0$ to a moderate value, the sum rate improves rapidly and 
then becomes nearly saturated around $\ell=6$ to $12$, indicating that a small number of demonstration pairs already provides effective anchors for query beamforming inference. When $\ell$ further increases, the performance slightly decreases, possibly because redundant or less relevant demonstrations 
dilute the attention to the most informative context samples. Based on this result, we set $\ell=5$ for the subsequent simulations, which achieves near-peak performance with a small context length. More importantly, the severe performance degradation at $\ell=0$, where the model only observes the query pilot token without any demonstration pair, confirms the necessity of context information and justifies the effectiveness of the ICL mechanism for pilot-based beamforming.

\begin{figure} [htbp]
    \centering
    \includegraphics[width=1.05\linewidth]{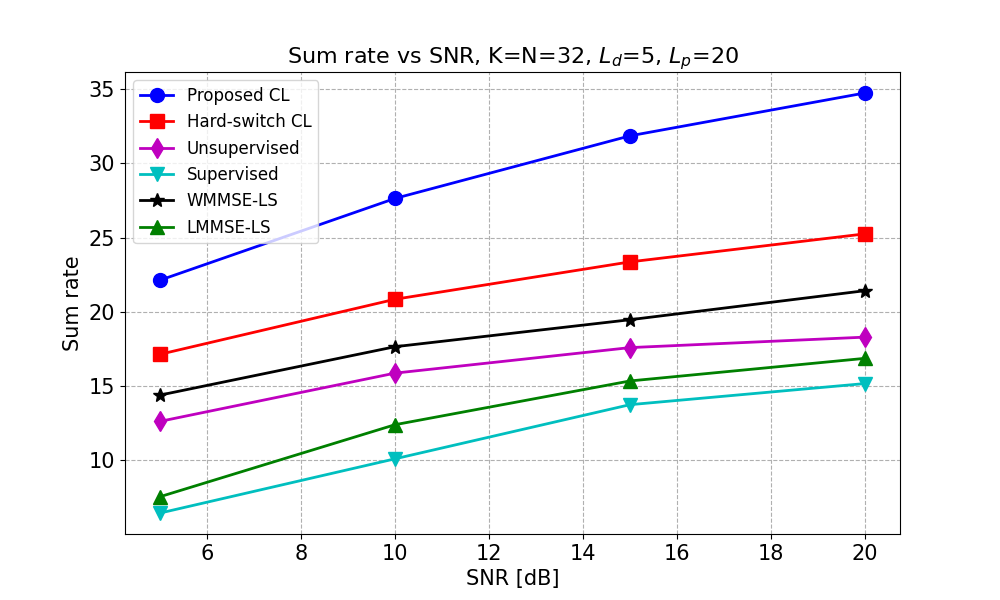}
    \caption{Comparison between the proposed CL scheme and other training methods.}
    \label{cl_ablation}
\end{figure}

We then demonstrate the performance of the \textbf{proposed CL scheme} by comparing it with the following learning-based variants: (a) an \emph{unsupervised baseline}, which trains the model solely with the sum-rate objective and removes the supervised warm-start stage in Sec.~\ref{warm_start_phase}; (b) a \emph{supervised baseline}, which trains the model only by imitating LMMSE beamformers and removes the CL stage in Sec.~\ref{self_evolve_cl}; and (c) \emph{hard-switch CL baseline}, which rapidly changes $r_u=0$ to $r_u=1$ in \eqref{hybrid_loss_obj}, while the proposed scheme gradually increases $r_u$ from $0$ to $1$ with step size $0.1$. All schemes are evaluated over \(\text{SNR}\in\{5,10,15,20\}\) dB. Fig.~\ref{cl_ablation} plots the average testing sum rate versus SNR. It is seen that the proposed CL scheme consistently outperforms all three learning-based variants, as well as the WMMSE-LS and LMMSE-LS baselines. The poor performance of the unsupervised baseline shows that directly optimizing the sum-rate objective from scratch is ineffective, confirming the necessity of supervised warm-start. In contrast, the supervised baseline is limited by LMMSE-label imitation and cannot achieve further task-oriented improvement. Although hard-switch CL improves over these two extremes, its clear gap from the proposed scheme shows that a smooth transition from label imitation to sum-rate maximization is crucial for stable and effective training.


\begin{figure} [htbp]
    \centering
    \includegraphics[width=1.05\linewidth]{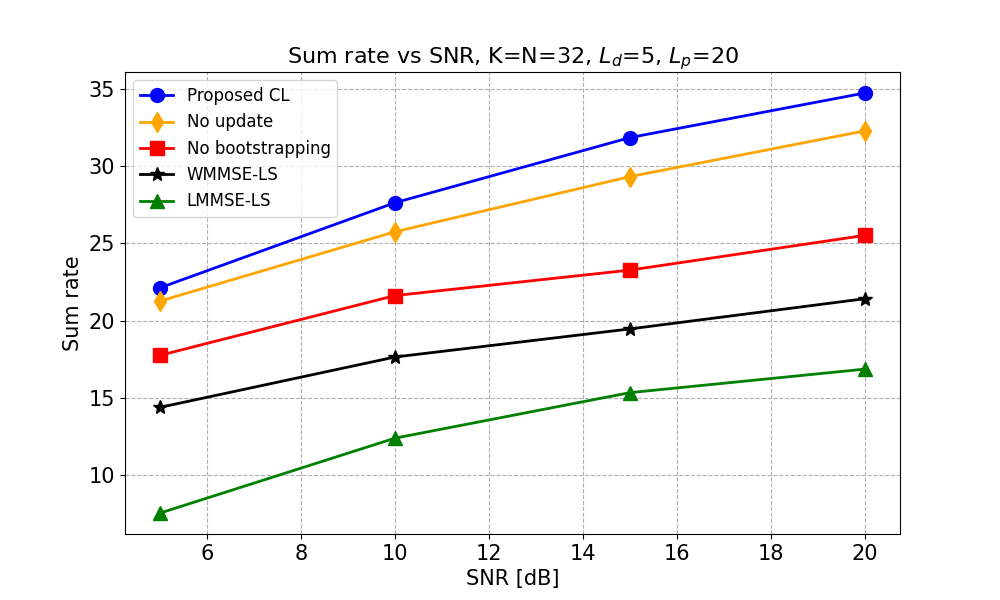}
    \caption{Comparison between the proposed context-bootstrapping scheme and other ICL schemes.}
    \label{bootstrap_ablation}
\end{figure}

Finally, we examine the effectiveness of the proposed \textbf{context bootstrapping} mechanism, a core component of the enhanced ICL scheme. Two variants are considered for comparison: (a) a \emph{No bootstrapping baseline}, which uses the same CL strategy but keeps each context dataset fixed as the initial LMMSE-labeled seed dataset throughout training; and (b) a \emph{No update baseline}, which admits newly generated samples into the context dataset but does not refine existing samples. All schemes are evaluated over \(\text{SNR}\in\{5,10,15,20\}\) dB. Fig.~\ref{bootstrap_ablation} plots the average testing sum rate versus SNR. The ``No bootstrapping'' baseline lags behind the proposed scheme, indicating that the initial LMMSE-labeled dataset alone cannot provide sufficiently informative contexts. The ``No update'' baseline improves over ``No bootstrapping'' baseline by expanding the context dataset, but its remaining gap to the proposed scheme shows that updating existing samples is also important. Therefore, the proposed bootstrapping mechanism benefits from both admitting high-quality fresh samples and continuously refining existing context samples.



\subsection{Overall Performance} \label{overall_result}

\begin{figure*} [htbp]
    \centering
    \includegraphics[width=1.02\linewidth]{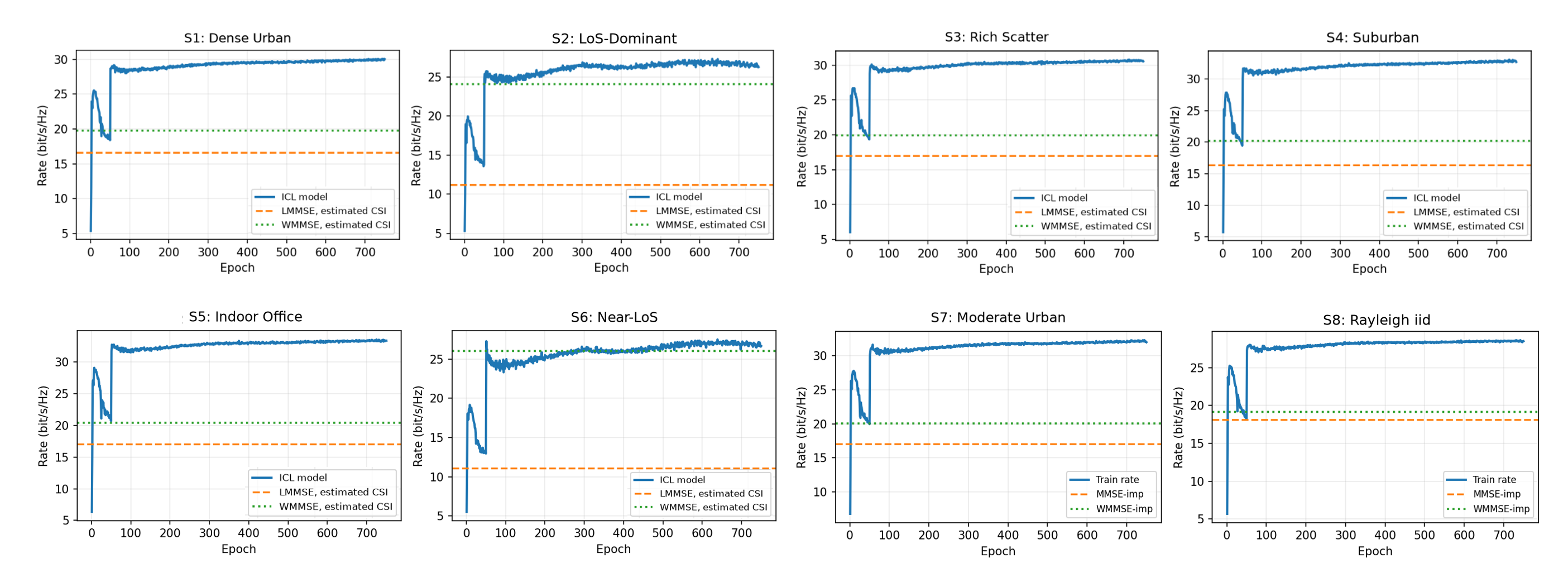}
    \caption{On-the-fly testing performance of the enhanced ICL beamforming scheme under multiple channel models.}
    \label{otf_conv_multi_scenario}
\end{figure*}

The overall performance of the proposed enhanced ICL beamforming scheme is now evaluated. We first examine its \textbf{convergence performance} across multiple channel models. 
For each channel model, a fixed testing set is generated before training and used throughout the entire training process, thereby enabling a consistent on-the-fly evaluation of the model performance. Fig.~\ref{otf_conv_multi_scenario} reports the test sum rate curves for eight representative channel models in Table~\ref{multi_scenarios}. As observed, the proposed scheme exhibits stable convergence behavior across all considered channel models and eventually outperforms both WMMSE-LS and LMMSE-LS baselines, demonstrating its excellent multi-model adaptability. Although a temporary degradation occurs when the training stage switches, the sum rate increases rapidly and then gradually improves as the context dataset is refined during training. Among all channel models, the relative performance gains in the ``LoS-dominant'' and ``Near-LoS'' channel models are smaller, mainly because that a dominated propagation component can increase the spatial correlation among user channels and reduce their separability, making interference suppression and pilot-to-beamformer learning more challenging. Nevertheless, the proposed scheme still achieves competitive or superior performance in these challenging scenarios, showing its robustness across heterogeneous channel environments.

We then show that the proposed ICL scheme is capable to perform a fast \textbf{online adaptation to an unseen} channel model, i.e., $s'\notin\mathcal{S}$ and the corresponding channel distribution $\bm{H} \sim p_{s'}$ is unknown. Instead of fine-tuning the trained ICL model, the BS constructs a small channel-specific context dataset at inference time: it collects pilot signals $\{\boldsymbol Y'_{j}\}_{j=1}^{M_0}$, obtains LS channel estimates $\bm{H}'_{j}=\bm{Y}'_{j} \bm{\Psi}^{\dagger}$, and computes the corresponding LMMSE beamformers $\bm{W}'^{\text{m}}_{j}(\bm{H}'_{j})$. The resulting context dataset is then given by $\mathcal{V}_{s'} = \{(\bm{H}'_{j},\bm{Y}'_{j},\bm{W}'^{\text{m}}_{j})\}_{j=1}^{M_0}$. Conditioned on $\mathcal{V}_{s'}$, the trained ICL model generates beamforming solutions for new channel model without online backpropagation or parameter updates. Specifically, Table~\ref{unseen_adapt} reports the test sum-rate performance under three additional channel models, beyond the eight basic channel models summarized in Table~\ref{multi_scenarios}. For each model, we consider two experimental settings. In the \emph{Seen} setting, the target channel model is included in the training model set $\mathcal{S}$; in the \emph{Unseen} setting, it is excluded from $\mathcal{S}$.
It shows that the proposed ICL scheme retains strong performance even when the target channel model is unseen during training. In all three models, the \emph{Unseen} setting outperforms both LMMSE-LS and WMMSE-LS. Compared with the corresponding \emph{Seen} setting, the sum-rate degradation remains moderate, despite the complete exclusion of the target channel distribution from offline training. This confirms the adaptation capability of the proposed ICL mechanism to previously unseen channel models.


\begin{table}[t]
    \centering
    \caption{Testing sum-rate performance when channel models are seen or unseen during training.}
    \label{unseen_adapt}
    \begin{tabular}{c|c|c|c|c}
        \hline
        Channel model & Unseen & Seen & LMMSE-LS & WMMSE-LS \\
        \hline
        Highway      & 26.586 & 29.244 & 15.589 & 20.880 \\
        Dense Indoor & 27.624 & 30.382 & 17.420 & 20.069 \\
        Rural LoS    & 25.871 & 28.435 & 10.490 & 25.477 \\
        \hline
    \end{tabular}
\end{table}

\begin{figure} [htbp]
    \centering
    \includegraphics[width=1.08\linewidth]{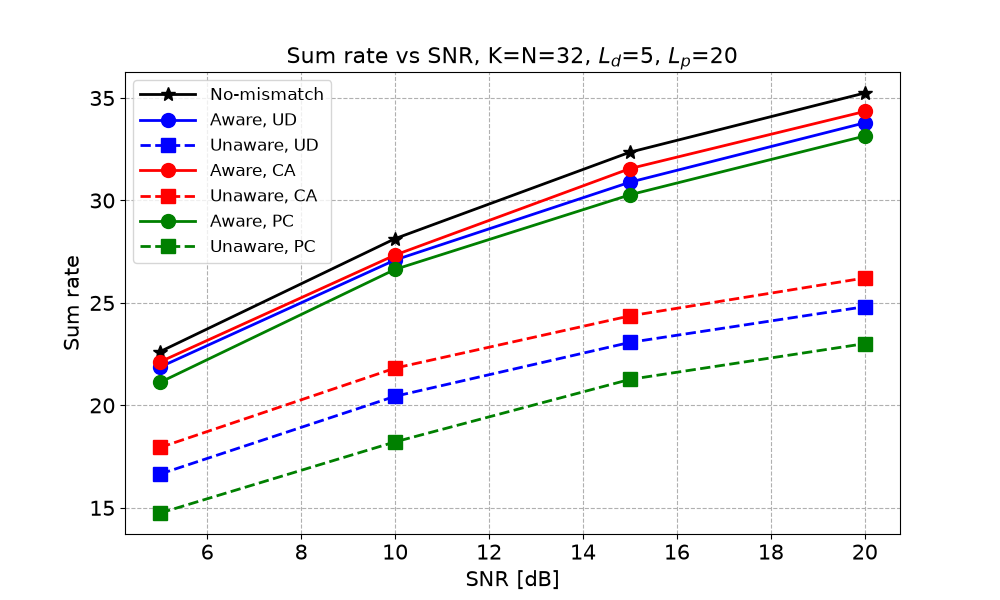}
    \caption{Performance of the mismatch-aware ICL schemes under different mismatch types. ``UD'', ``CA'' and ``PC'' represent the uplink-downlink mismatch, channel-aging mismatch, and pilot-contamination mismatch, respectively. ``Aware'' and ``Unaware'' represent the corresponding mismatch-aware and mismatch-unaware ICL schemes.}
    \label{mismatch_performance}
\end{figure}

Next, we evaluate the proposed scheme under the presence of model mismatches. The ICL network is trained over all channel models in Table~\ref{multi_scenarios} and three mismatch types introduced in Sec.~\ref{exten_model_mismatch}, while the testing results are reported on the ``Dense Urban'' channel model with each type of mismatch.   
Three ICL schemes are provided for comparison: (a) the proposed \emph{mismatch-aware ICL} scheme, where pilot signals are generated from the mismatched channels but beamformers are evaluated on the original channels; (b) \emph{no-mismatch benchmark} that serves as an upper bound, where pilot signal and the beamformer evaluation are both associated with the original channels; and (c) the \emph{mismatch-unaware ICL} scheme, where pilot signals and the beamformer evaluation are both associated with the mismatched channels. Fig.~\ref{mismatch_performance} plots the average testing sum rate versus SNR under different mismatch types and ICL schemes. It shows that all mismatch-unaware ICL baselines suffer significant performance degradations, due to deviated solutions in the context. In contrast, the proposed mismatch-aware ICL schemes substantially improve the sum rate and remain close to the common upper bound (no-mismatch benchmark). This demonstrates that the proposed ICL scheme can effectively absorb the corresponding mismatches into the context constructions, without explicit channel calibration operations.

\begin{figure} [htbp]
    \centering
    \includegraphics[width=1.07\linewidth]{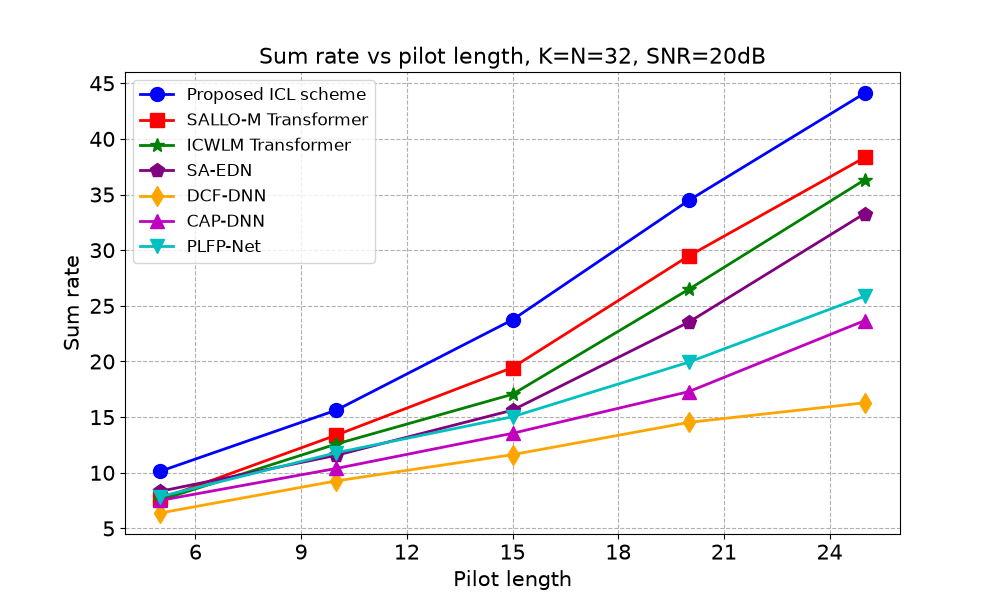}
    \caption{Comparison between the proposed scheme and baselines under varying pilot lengths $L_p$.}
    \label{baseline_comparison}
\end{figure}

We finally compare the proposed scheme with existing learning-based beamforming schemes. In addition to the WMMSE-LS and LMMSE-LS baselines introduced above, the following methods are considered:
\begin{itemize}
    \item \textbf{ICWLM}: a Transformer-based supervised ICL model for multi-task physical-layer optimization, originally using CSI-beamformer demonstration pairs~\cite{wen2026icwlm};
    \item \textbf{SALLO-M}: a semi-amortized L2O-based Transformer optimizer for beamforming based on perfect CSI~\cite{zhang2025sallom};
    \item \textbf{SA-EDN}: a semi-amortized encoder-decoder network for
    beamforming based on estimated CSI~\cite{zhang2026semi};
    \item \textbf{DCF-DNN}: an end-to-end learning framework that maps received pilots
    to distributed feedback and subsequently to beamformer~\cite{sohrabi2021dl_feedback};
    \item \textbf{CAP-DNN}: an end-to-end learning framework with channel-adaptive pilot generation and direct beamformer prediction~\cite{park2024e2e_tdd};
    \item \textbf{PLFP-Net}: a WMMSE-guided framework that jointly learns
    pilot transmission, limited feedback, and beamforming~\cite{jang2022joint_pilot_feedback}.
\end{itemize}
For a fair comparison, all methods are evaluated under the same system dimensions $N=K=32$, transmit SNR$=20$dB, power constraint $P=1$, while pilot length $L_p$ is increased from $5$ to $25$ with step size $5$. No mismatches are considered in this experiment. Note that perfect CSI is not available --- whenever CSI is required by a baseline, it is replaced by the LS estimate $\hat{\bm H}=\bm Y\bm\Psi^\dagger$, obtained from available pilot signal $\bm{Y}$. 
Accordingly, for the ``SALLO-M'' baseline, the auxiliary variable and beamformer are initialized by $\hat{\bm H}$ and its LMMSE beamformer $\bm{W}^{\text{m}}(\hat{\bm H})$, respectively. 
The ``SA-EDN'' baseline likewise uses $\hat{\bm H}$ as its input. For ``ICWLM'' baseline, we maintain its supervised training procedure and replace each CSI $\bm{H}$ by the corresponding pilot signal $\bm{Y}$ in the demonstration pairs, as introduced in Sec.~\ref{prob_state}, to adapt to the pilot-based setting. For ``DCF-DNN'' and ``PLFP-Net'' baselines, their native pilot-feedback-beamformer pipelines are retained under the same pilot settings. The ``CAP-DNN'' baseline preserves its channel-adaptive pilot design, but the required CSI is restricted to the LS estimate $\hat{\bm H}$. Importantly, our proposed scheme and ``ICWLM'' baseline are able to be trained over the same set of channel models listed in Table~\ref{multi_scenarios} and tested on the ``Dense Urban'' channel model, while the remaining baselines lack the multi-model adaptability and hence are trained and tested only on the ``Dense Urban'' channel model. All learning-based methods use the same testing channel samples. Fig.~\ref{baseline_comparison} plots the average testing sum rate versus the
pilot length $L_p$. It is seen that all schemes benefit from increasing $L_p$, and the proposed scheme consistently achieves the highest sum rate over the other schemes. It also shows that all Transformer-based schemes outperform the other DNN-based schemes, suggesting a superior learning capability of the Transformer model. Among those Transformer-based schemes, the advantage of the proposed scheme over ``SALLO-M'' and ``ICWLM'' baselines indicates that both the in-context structure and enhanced training strategies contribute to the final performance.

\section{Conclusion}

We proposed a self-evolving ICL framework for direct pilot-to-beamformer design in MU-MISO systems. By integrating pilot and beamformer EDNs with a shared ICL Transformer, the proposed framework directly generates beamformers from noisy, limited-length pilots without explicit channel estimation. Its CL-based training strategy together with the self-evolving context bootstrapping mechanism improve the convergence behavior and substantially reduce the reliance on near-optimal labeled data. Through intelligent model-specific context constructions, a single shared network can adapt to diverse channel models and mismatches, without parameter updates or explicit channel calibrations. Extensive simulations demonstrate high-quality adaptations to both seen and unseen channel models, robust performance under model mismatches, and consistent gains over existing pilot-based beamforming methods. Future work will extend the proposed framework to broader physical-layer optimization tasks, and evaluate its performance using measurement-based channels and over-the-air experiments.

\bibliographystyle{IEEEtran}
\bibliography{IEEEabrv,bibfile}

\end{document}